\documentclass[sn-mathphys,Numbered]{sn-jnl}
\usepackage{graphicx} 
\usepackage{svg}
\usepackage{placeins}
\usepackage{xltabular}
\usepackage{makecell}
\usepackage{bbm}
\usepackage{amsfonts}
\usepackage{xcolor}
\usepackage{siunitx}
\usepackage{arydshln}
\usepackage{setspace} 
\usepackage{geometry} 
\usepackage{caption}
\usepackage{booktabs}
\usepackage{subcaption} 
\usepackage{algorithm}
\usepackage{algpseudocode}
\usepackage{amsmath} 
\usepackage{multirow} 
\usepackage{amssymb} 
\usepackage{cuted}
\usepackage[nottoc,notlot,notlof]{tocbibind} 
\usepackage{float}
\usepackage{lscape}
\usepackage{adjustbox}

\begin{document}

\title{Layer-wise Model Merging for Unsupervised Domain Adaptation in Segmentation Tasks}


\author*[1]{\fnm{Roberto} \sur{Alcover-Couso}}\email{roberto.alcover@uam.es}

\author[1]{\fnm{Juan C.} \sur{ SanMiguel}}\email{juancarlos.sanmiguel@uam.es}

\author[1]{\fnm{Marcos} \sur{ Escudero-Viñolo}}\email{marcos.escudero@uam.es}

\author[1]{\fnm{Jose M} \sur{Martínez}}\email{josem.martinez@uam.es}

\affil*[1]{\orgname{Video Processing and Understanding Lab}, \orgdiv{Universidad Aut\'{o}noma de Madrid (UAM)}, \orgaddress{\city{Madrid}, \postcode{28049}, \country{Spain}}}

\abstract{

Merging parameters of multiple models has resurfaced as an effective strategy to enhance task performance and robustness, but prior work is limited by the high costs of ensemble creation and inference.
In this paper, we leverage the abundance of freely accessible trained models to introduce a cost-free approach to model merging. It focuses on a layer-wise integration of merged models, aiming to maintain the distinctiveness of the task-specific final layers while unifying the initial layers, which are primarily associated with feature extraction. This approach ensures parameter consistency across all layers, essential for boosting performance. Moreover, it facilitates seamless integration of knowledge, enabling effective merging of models from different datasets and tasks. 
Specifically, we investigate its applicability in Unsupervised Domain Adaptation (UDA), an unexplored area for model merging, for Semantic and Panoptic Segmentation.
Experimental results demonstrate substantial UDA improvements without additional costs for merging same-architecture models from distinct datasets ($\uparrow 2.6\%$ mIoU) and different-architecture models with a shared backbone ($\uparrow 6.8\%$ mIoU). Furthermore, merging Semantic and Panoptic Segmentation models increases mPQ by $\uparrow$ 7\%. These findings are validated across a wide variety of UDA strategies, architectures, and datasets. 

}
\maketitle
\section{Introduction}

Unsupervised domain adaptation (UDA) aims to train models from a source labeled domain that can generalize to unlabeled target domains, key for tasks with limited annotated data due to the capture difficulty or the high annotation costs (e.g., semantic or panoptic segmentation \cite{Cordts2016Cityscapes}). Thus, automatically-annotated synthetic datasets emerge as attractive alternative for the source domain for segmentation tasks \cite{edaps,hoyer2022daformer,hoyer2022hrda,hoyer2023mic,Marcos-Manchon_2024_CVPR}.  While UDA has shown remarkable success in various Computer Vision tasks, UDA for segmentation tasks often faces unstable training \cite{PCCL,vu2019advent,Wang2020,Chen_2019_ICCV,8578878} and underperforms supervised models  \cite{alcovercouso2023soft, Xie2021SegFormerSA,YuanCW20,Diana-Albelda_2024_CVPR}. In this situation, teacher-student distillation has emerged as a standard technique for enhancing UDA training stability \cite{hoyer2022daformer, edaps, Chao_2021_CVPR,s23020621}, albeit with increased costs due to inferencing the teacher while training and inferencing the student \cite{s23020621, Chao_2021_CVPR}. 

Model merging arises as a promising technique to overcome the teacher-student computational overhead, as combining parameters directly only requires a single model inference \cite{ModelAverage92, tsypkin1971adaptation, wortsman2021robust, Tarvainen2017MeanTA}. 
Early research on model merging focused on learning convergence\cite{ModelAverage92, tsypkin1971adaptation}, but declined over time with newer training and regularization methods. Later, the rise of large language models and self-supervised learning  have reignited interest in model merging \cite{ilharco2023editing, pmlr-v162-wortsman22a, NEURIPS2022_70c26937}, as it provides a training-free mechanism to leverage knowledge from different sources. Recent methods apply model merging mostly to different checkpoints obtained during the same training process, eliminating additional computation costs for ensemble training and inference \cite{61aa9e9cc965421e82d7b7042c61abc8,ilharco2023editing, 10.5555/3524938.3525243,yamada2023revisiting}. However, these methods use models close in the parameter space, such as ones obtained from different seeds \cite{Tarvainen2017MeanTA,NEURIPS2022_70c26937} or different fine-tuning versions \cite{zhou2023permutation,Pena_2023_CVPR,ainsworth2023git, pmlr-v162-wortsman22a}, as merging models with misaligned layers typically leads to a notable performance decline \cite{LinearUnsuited,Pena_2023_CVPR,yamada2023revisiting,ainsworth2023git}. Therefore, current methods disregard to explore the possible benefits of combining the plethora of existing models trained using different methods and tasks. Additionally, to the best of our knowledge, model merging methods have not been explored and benchmarked for UDA \cite{li2023deep,edaps,hoyer2022daformer,hoyer2022hrda,hoyer2023mic}. 

\begin{figure}[tp]
    \centering
    \begin{subfigure}[b]{0.5\linewidth}
        \includegraphics[width=\linewidth]{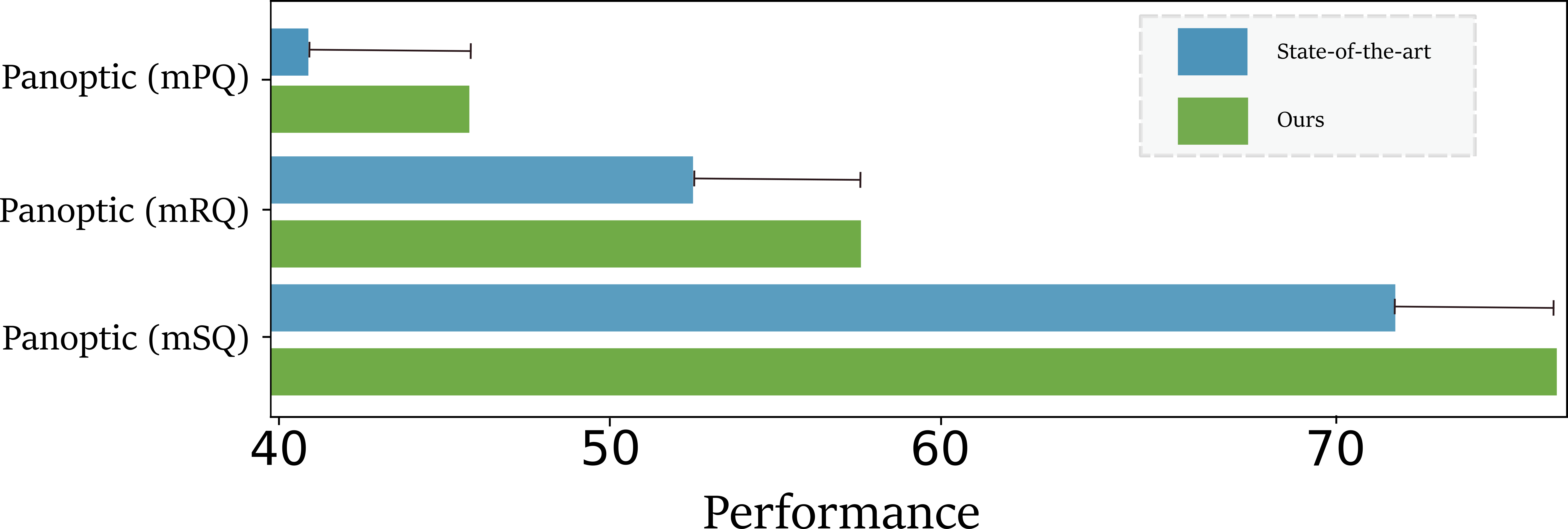}
        \caption{Panoptic segmentation.}
    \end{subfigure}\hfill
    \begin{subfigure}[b]{0.45\linewidth}
        \includegraphics[trim=0 0 0 148,clip,width=\linewidth]{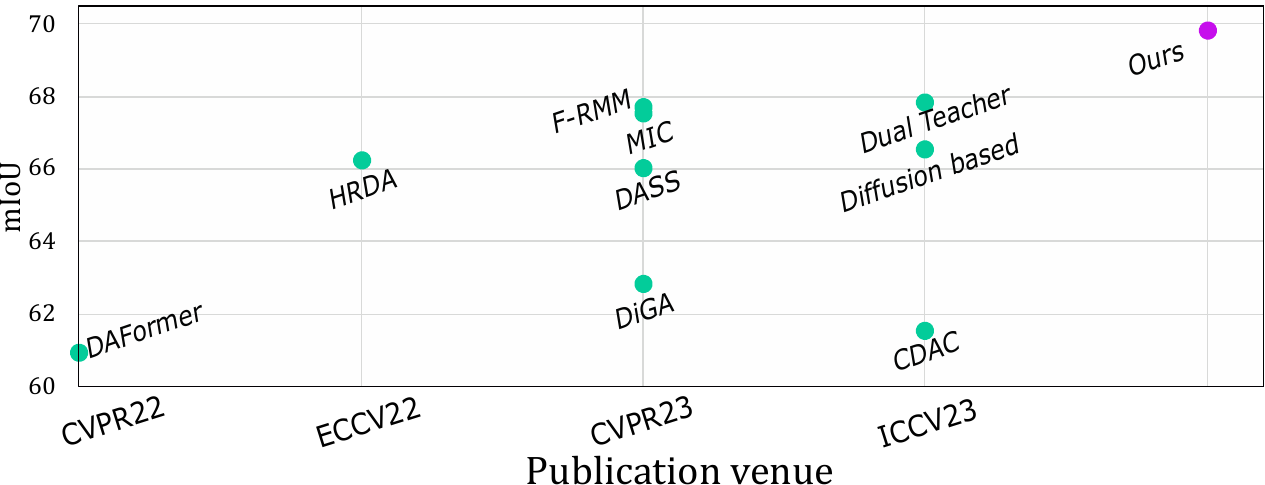}
        \caption{Semantic segmentation.}
    \end{subfigure}
    \caption{Performance comparison for the UDA Synthia-to-Cityscapes setup. Most state-of-the-art UDA methods build up from DAFormer \cite{hoyer2022daformer} by adding computationally expensive blocks. In contrast, we leverage the obtained checkpoints during training to improve performance without incurring any associated training or inference overhead.} 
    \label{fig:Fig1}
    \vspace{-5mm}
\end{figure}

To overcome the above-mentioned deficiencies, we propose a novel layer-wise merging method that exploits the specific strengths of deeper layers closer to the classification output and the broader generalization capabilities of shallower layers, making it capable of effectively merging models derived from different datasets and tasks. Moreover, we argue that the combination of model weights in the context of UDA has far more applications than the limited scope in which it is currently being employed (i.e., teacher-student framework) such as combining models from different tasks.
To evaluate our proposal for Semantic and Panoptic Segmentation, we set an extensive baseline by experimenting with various model merging methods based on the same or different datasets and architectures. Our experiments also address two challenges. The first one aims to incorporate high-performing yet computationally expensive models into faster but less accurate ones. Our proposal increases speed by 3.25 times with performance increases up to +6.8\% in mIoU. The second challenge is for combining models from easier tasks into harder tasks. Our proposal presents performance gains of +3.4\% in mIoU and +7\% in mPQ compared to previous state-of-the-art UDA methods for respectively Semantic and Panoptic Segmentation (see Figure \ref{fig:Fig1}). Our results are validated across various UDA synthetic-to-real setups for multiple strategies (including adversarial, self-training, and entropy minimization), architectures (convolutional and transformer based) and datasets (Cityscapes and Mapillary). 


Our main contributions can be summarized as follows:
\newline\textbf{(1)} Cost-free model merging by introducing a novel method to merge existing available pretrained models without requiring additional training costs.
\newline\textbf{(2)} Cross-task model merging by applying model merging to combine models designed for different tasks with shared parameters.
\newline\textbf{(3)} Extensive benchmarking of model merging for Unsupervised Domain Adaptation, focusing on multiple models, strategies and datasets.


\section{Related Work}
\textbf{Training strategies for Unsupervised Domain Adaptation (UDA).}
UDA has been widely explored for semantic \cite{PCCL,hoyer2022daformer, hoyer2022hrda,VC1}, and panoptic segmentation \cite{edaps,Huang_2021_CVPR,10203383}. Methods based on \emph{Deep Learning} focus on minimizing source-target differences using statistical metrics like maximum mean discrepancy \cite{10.1145/3474085.3475186, wmmd, 8099590}, correlation alignment \cite{10.1007/978-3-319-49409-8_35,coral,coral-lda}, or entropy minimization \cite{vu2019advent,Chen_2019_ICCV,NIPS2004_96f2b50b}. Others employ \emph{Adversarial training} by learning a domain discriminator within a Generative Adversarial Network (GAN) framework \cite{goodfellow2014generative} to enforce domain-invariant inputs \cite{Gong2018DLOWDF,pmlr-v80-hoffman18a,VC2}, features \cite{hoffman2016fcns,long2018conditional,VC1,VC3}, or outputs \cite{vu2019advent, Wang2020, 8578878}. Lastly, \emph{self-training} generates pseudo-labels \cite{Lee2013PseudoLabelT} for the target domain based on predictions obtained using confidence thresholds \cite{mei2020instance,8578498, Marcos-Manchon_2024_CVPR} or pseudo-label prototypes \cite{8954306,zhang2021prototypical}. More recent methods \textit{based on Transformers} tackle noisy training and concept drifting \cite{CascanteBonilla2020CurriculumLR, Sammut2010} by incorporating consistency regularization \cite{10.5555/3157096.3157227, 10.5555/3495724.3495775} to ensure uniformity across diverse data augmentations \cite{Araslanov:2021:DASAC,melaskyriazi2021pixmatch}, different crops \cite{hoyer2022hrda, lai2021cac}, domain-mixup \cite{Tranheden2020DACSDA,Alcover-Couso_2023_ICCV, hoyer2022hrda,10.1007/s11263-023-01799-6} or curriculum learning \cite{PCCL, MTAP}. Currently, enhancing robustness is dominated by the teacher-student framework \cite{Tarvainen2017MeanTA}, where the teacher model is updated via an exponential moving average of the student weights \cite{hoyer2022daformer, edaps, hoyer2022hrda,hoyer2023mic}. Albeit effective, this framework incurs in additional computational costs which limit its scalability and practicality for resource-constrained environments.

\textbf{Leveraging models knowledge.}
Averaging model weights initially boosted neural network training by addressing slow algorithms with poor convergence rates \cite{61aa9e9cc965421e82d7b7042c61abc8,ModelAverage92, tsypkin1971adaptation,Tarvainen2017MeanTA}. However, as training methodologies evolved, averaging efficacy diminished for linear problems \cite{LinearUnsuited}, prompting a pivot to output ensembles and fine-tuning strategies. \textit{Output ensembles} combine predictions from multiple models \cite{MOHAMMED2023757,HASHEM1997599}. \textit{Fine-tuning techniques} adjust pre-trained model weights with minimal training for better task adaptation, enhancing transferability and generalization \cite{Houlsby2019ParameterEfficientTL, FinetuningTC, zhang2021revisiting, hu2022lora}. Research indicates that deeper model layers contain task-specific filters, while shallow layers have task-agnostic filters, suggesting minimal tuning of the latter for best performance \cite{FinetuningTC, zhang2021revisiting,ilharco2023editing, 10.5555/3524938.3525243}. Thus, a layer-specific model merging approach is essential for effective combination.

\begin{table*}[tp]
\centering
    \setlength\tabcolsep{7pt}
    \resizebox{1\linewidth}{!}{%
    \begin{tabular}{l c c c c c c}
        Method & Weight- & Data & Different & Different & Ensemble & Inference \\
               & level & Free & Dataset & Head & Cost & Cost \\\toprule
         Isotropic \cite{ ModelAverage92,pmlr-v162-wortsman22a, wortsman2021robust}&\checkmark &\checkmark  &$\times$&$\times$& O(1) & O(1)\\
         Fisher weighted \cite{NEURIPS2022_70c26937} &\checkmark&$\times$&$\times$&$\times$& O(N)& O(1) \\
         Weight permutation \cite{ainsworth2023git,Pena_2023_CVPR} &\checkmark&$\times$&$\times$&$\times$& O(N)& O(1) \\
         Output ensemble \cite{ HASHEM1997599}&$\times$ & \checkmark&\checkmark& \checkmark& O(1) & O(M) \\
         Layer-wise (Ours) &\checkmark&\checkmark&\checkmark &\checkmark& O(1) & O(1)\\\bottomrule
    \end{tabular}}
    \caption{Model merging methods comparison. Data free denotes that do not require any image data. (Key. M: Number of models merged. N: Number of data samples.)}
    \label{tab:WeightingComp}
    \vspace{-5mm}
\end{table*}

\textbf{Merging models weights.}
Current merging methods can be applied during training \cite{ModelAverage92,pmlr-v162-wortsman22a, hoyer2022daformer} or post-training \cite{NEURIPS2022_70c26937, yamada2023revisiting, ainsworth2023git,Pena_2023_CVPR}. The standard isotropic merging provides a straight-forward way to merge the models by taking their average \cite{NEURIPS2022_70c26937, ModelAverage92,pmlr-v162-wortsman22a, wortsman2021robust}. However, methods employing a performance-based weighted average of model parameters outperformed isotropic merging in different tasks \cite{wortsman2021robust,NEURIPS2022_70c26937}.
Alternatively, some proposals aim to assign layer-level weights by employing Fisher information of the model’s parameters on the training data \cite{ NEURIPS2022_70c26937}. However, this merging fails to combine batch normalization (BN) mean and standard deviation parameters, as they are not computed through gradient descent. Moreover, this merging strategy disregards the architectural structure, potentially leading to ineffective contributions from different models across the layers. Moreover, permutation of weights methods \cite{yamada2023revisiting, ainsworth2023git,Pena_2023_CVPR} attempt to align model weights across different initializations to improve the weight combination. This strategy employs the loss landscape on the different models across the dataset to learn a permutation over the weights of both models so that the combined filters encapsulate similar information \cite{ainsworth2023git}. Nevertheless, this permutation is not suited for merging models from different datasets as their loss landscapes are expected to differ \cite{yamada2023revisiting}. In the context of UDA, weight permutation methods are not suited as the loss landscapes differ from source to target domains. Furthermore, as models are typically initialized with ImageNet pre-trained weights, initial layers of the models are expected to be aligned \cite{ilharco2023editing, Tarvainen2017MeanTA}.
Table \ref{tab:WeightingComp} compares our merging to alternative model merging methods. Note that output ensemble combines the predictions of different models. Therefore, at inference each of the  models have to generate a prediction, hence, drastically increasing inference costs as it is not a weight-level model merging method.

\textbf{Model merging in UDA.}
Model output ensemble for UDA is a standard mechanism to improve performance across different visual tasks \cite{PIVA2023103745,Chao_2021_CVPR,9157742,s23020621}. These ensembles can be employed directly  on validation \cite{PIVA2023103745,9157742} or training as a knowledge distillation mechanism to generate reliable pseudo-labels to train a student model \cite{9157742,s23020621}. To improve knowledge distillation efficiency, current advancements \cite{hoyer2022daformer,hoyer2022hrda} define the teacher model weights as a running mean of the student model weights \cite{Tarvainen2017MeanTA}.  However, the use of model merging is limited as the student model does not directly benefit from it. To the best of our knowledge, we are the first study that has quantified the advantages of model merging in UDA, proposed alternatives to isotropic merging for UDA and explored the merging of student models across architectures and tasks.

\section{Layer-wise model merging}
Our proposal relies on previous research for fine-tuning \cite{FinetuningTC, zhang2021revisiting,ilharco2023editing, 10.5555/3524938.3525243}. By comparing the initial and the resulting weights after fine-tuning to a specific task, it is suggested that deeper layers in the model encapsulate task-specific filters. Meanwhile, shallow layers represent task-agnostic filters which should be less tuned to obtain optimal performance. Our work follows this hypothesis to define a layer-wise parameter merging, having early layers merged in a uniform manner, as they encode general knowledge and their smoothing should provide more robust features \cite{61aa9e9cc965421e82d7b7042c61abc8, ModelAverage92, tsypkin1971adaptation,Tarvainen2017MeanTA}. While preserving last layers so that the task-specific knowledge is retained after merging \cite{FinetuningTC, zhang2021revisiting,ilharco2023editing, 10.5555/3524938.3525243}.

The following subsections detail the assumptions of our proposal, investigate this hypothesis for the UDA setup, and provide the definition of how model merging is conducted.

\subsection{Preliminaries}

In the context of this work, we define an architecture as the specific arrangement of layers and their connections within a neural network. A model refers to this architecture after it has been trained on a particular dataset using a specific training regimen. We further distinguish between the backbone and the classification head of the model: the backbone consists of the initial and intermediate layers responsible for feature extraction, while the classification head comprises the final layers designed for task-specific purposes, tailored to a specific set of classes and tasks.

Let $M$ be the number of trained models, each one with its respective parameters defined by $\theta_1,\hdots,\theta_i,\hdots,\theta_M$. We consider that each model can be decomposed into a number of layers, where $\theta_i^{(j)}$ are the parameters of the $ith$ model and the $jth$ layer. Initial and intermediate layers form the backbone, whereas the model's last layers correspond to the classification head, tailored to the specific task.

Following the assumption prevalent in related work \cite{ModelAverage92,pmlr-v162-wortsman22a, wortsman2021robust,NEURIPS2022_70c26937, ainsworth2023git,Pena_2023_CVPR, HASHEM1997599}, we presume access to the performance metrics of models prior to initiating the merging process. Furthermore, our approach assumes the availability of pre-trained models, aligning with contemporary practices centered on foundational models where extensive pre-training on large datasets provides a robust starting point for adapting to various tasks \cite{awais2023foundational}. Lastly, we assume all models undergoing merging have undergone the same pre-training for the shared parameters, ensuring their future alignment \cite{NEURIPS2022_70c26937}. Despite variations from different training schemes, the models' shared parameters should remain relatively close within the parameter space, similarly to \cite{NEURIPS2022_70c26937}.

\subsection{Exploring heterogeneity of models in UDA}
\begin{figure}[tp]
    \centering
    \includegraphics[width=.57\linewidth]{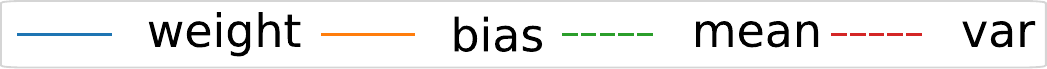}
    \begin{subfigure}[b]{0.46\linewidth}
        \vspace{4mm}
        \includegraphics[width=\linewidth]{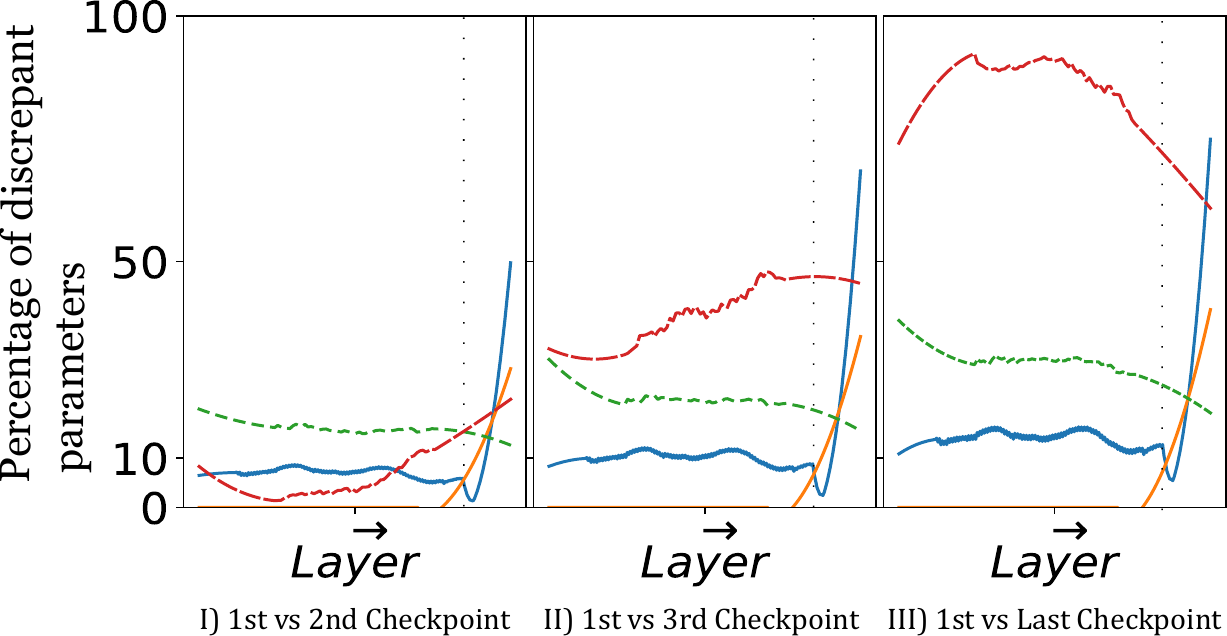}
        \caption{Checkpoint training scenario.}
        \label{fig:checkpoint_advancement_a}
    \end{subfigure}
    \begin{subfigure}[b]{0.46\linewidth}
        \includegraphics[width=\linewidth]{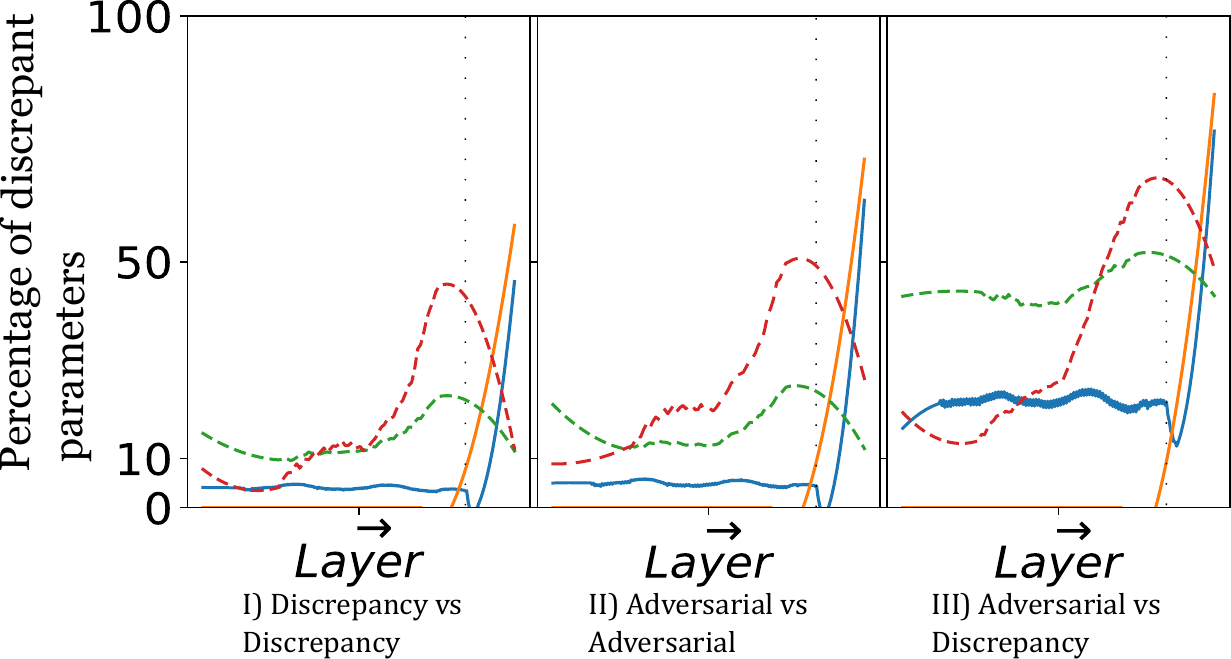}
        \caption{Heterogeneous training scenario.}\vspace{-2mm}
        \label{fig:checkpoint_advancement_b}
    \end{subfigure}
    
    \caption{Layer-wise discrepancy between models under two scenarios. The \textit{Checkpoint training scenario} assumes an available training process with four checkpoints in total, and compares the first checkpoint against the other three. The \textit{Heterogeneous training scenario} considers the availability of two models trained with different UDA strategies (discrepancy and adversarial, as defined by \cite{vu2019advent}), and compares the best trained model with each method. The parameters considered include both convolutional (\textit{weights} and \textit{biases}) and batch normalization ones (\textit{mean} and \textit{variance}). All results are based on training the DeeplabV2 architecture \cite{7913730} for the GTA-to-Cityscapes setup. For visualization purposes, we have included a dotted line indicating the last backbone parameter and first segmentation head parameter.}
    
    \label{fig:checkpoint_advancement}
\end{figure}

We analyze model variations in two contexts: consistent models throughout training epochs (Checkpoint training scenario) and different models shaped by varied strategies (Heterogeneous training scenario). Figure \ref{fig:checkpoint_advancement} illustrates the layer-by-layer differences in model parameters, including both convolutional (\textit{weights} and \textit{biases}) and batch-normalization parameters (\textit{mean} and \textit{variance}). The discrepancy is measured as the number of parameters with relative absolute Euclidean distance above a certain threshold $\tau$: $||\theta_{m}^{(j)} - \theta_{m+z}^{(j)}|| >= \frac{||\theta_m^{(j)}||}{\tau}$.

Upon analyzing the differences between checkpoints (see Figure \ref{fig:checkpoint_advancement_a}), we observe various behaviours in the parameters analyzed. Globally, convolutional parameters show significant changes mainly in the classification head (i.e., last layers). On the other hand, \textit{weight} parameters undergo slight changes throughout the training process in backbone (i.e., initial and middle layers). This finding aligns with the training process, which assigns a learning rate ten times larger to the classification head compared to the backbone \cite{Wang2020,vu2019advent,8578878,Chen_2019_ICCV}. Thus, it is anticipated that head parameters would undergo substantial changes compared to backbone parameters, reflecting their adaptation to the specific task.

In contrast, the batch-normalization parameters (mean and variance), undergo substantial transformations as the training progresses in the different checkpoints. These changes are more pronounced and do not appear to align with the small changes observed in the convolutional layers. This observation suggests that these parameters in UDA are relatively unaligned with the learned parameters. Similarly, discrepancies across UDA methods  (see Figure \ref{fig:checkpoint_advancement_b}) are found at the last layers of the models and in batch-normalization parameters. These findings suggest that initial layers of the model should be combined in a different manner than final layers. 

The findings in Figure \ref{fig:checkpoint_advancement} indicate that effective model merging should account for the different functions of the backbone and head. The backbone's general parameters need thoughtful merging with head parameters to maintain feature extraction capabilities across tasks. Merging backbone and head parameters separately could yield suboptimal outcomes by disrupting the consistency between their parameter spaces.

Based on these findings, we propose a simple method aligned with the hypothesis that final layers should be preserved while maintaining a uniform weight merging of initial layers. Moreover, weight merging must have consistency between the backbone and the head parameters. 



\subsection{How is model merge conducted}
We consider that the $M$ models can be trained with different training strategies or multiple checkpoints which share a set of parameters $\{\theta_i^{(j)}\}_{j=1}^{N_p}$. In order to perform the merging, we first select an anchor model $\theta_{i^{anchor}}$, from the $M$ models to be combined, this anchor model will be the main contributor of knowledge for the merging. Then we propose to combine them by a layer-level weighting in order to get the merged layer $\theta^{(j)}_*$: 
\begin{equation}
\theta^{(j)}_* =
\begin{cases}
     \sum_i^MH_i^{(j)} \theta_i^{(j)} &\text{if }  j \in[1,…,N_p] \\
    \theta^{(j)}_* = \theta_{i^{anchor}}^{(j)} &  \text{otherwise}.
\end{cases}
\end{equation}
where $\theta_i^{(j)}$ represents the $j$-th layer of model $\theta_i$, and $H_i^{(j)}$ is the corresponding layer-level weight. Notably, the layer-level weights for each layer are normalized such that $\sum_i^M H_i^{(j)} = 1$, and the weight of the anchor model $i^{anchor}$ is greater than the one assigned to the rest: $H^{(j)}_{i^{anchor}} > H^{(j)}_{i} \text{ } \forall  j,i \text{ }|\text{ }i \neq i^{anchor}$.

We propose to uniformly merge the initial layers and gradually decrease the weight given to final layers for the non-anchor models, while increasing the weight given to final layers of the anchor model, denoted as $\theta_{i^{anchor}}$.
Specifically, $H_i^{(j)}$ is  defined as: $H_i^{(j)} = \frac{N_p-j}{N_pM} \text{ } \forall i \text{ }|\text{ } i\neq i^{anchor}$  and $j\in[1,…,N_p]$ are the indexes of shared parameters. If the merged models share all the parameters, then $N_p = card(\theta)$.  Notably, this merging allows for the merging of models with different number of classes, unlike alternative merging methods restricted to the same architecture for the classification heads. Thus, being our proposal the only viable solution for merging models trained on different tasks and datasets (see Table \ref{tab:WeightingComp}). 

\paragraph{How is inference conducted with merged models.}
As depicted by Table \ref{tab:WeightingComp} weight-level merging techniques do not have additional inference costs. This is because the models are computed at a checkpoint level and stored to be loaded as any other model checkpoint. Therefore,  for inference, output ensemble models employ: $\sum_i^M \frac{f(\cdot,  \theta_i)}{M}$ while our models employ: $f(\cdot,  \theta_*)$.

\section{Experimental Exploration}
\subsection{Experimental setup}
\paragraph{Image segmentation}
\textbf{Datasets.}
We employ popular UDA datasets for semantic and panoptic segmentation: GTA\cite{Richter_2016_ECCV} and Synthia \cite{Ros2016} as source datasets; Cityscapes \cite{Cordts2016Cityscapes} and Mapilliary \cite{MVD2017} as target datasets. 
\textbf{GTA} is a synthetic dataset with urban scenes composed of 25K images from the video-game Grand Theft Auto V, which shares 19 classes with Cityscapes. 
\textbf{Synthia} is a collection of different synthetic urban scenes datasets, we pick its subset SYNTHIA-RAND-CITYSCAPES which  is composed of 9.5K images sharing 16 classes with Cityscapes.
\textbf{Cityscapes} is a real dataset with urban scenes generated by filming with a camera inside of a car while driving. It consists of 3K images for training and 0.5K images for validation, with 19 annotated classes.
The \textbf{Mapillary} dataset is another real urban scene dataset, which comprises 18K and 2K images for training and validation, with 152 annotated classes.

\textbf{Evaluation metrics.}
For semantic segmentation, we employ mean per-class intersection over union (mIoU) \cite{Everingham10thepascal}, between the model prediction and the ground-truth label. IoU measures at pixel-level the relationship between True Positives (TP), False Positives (FP) and False Negatives (FN): $IoU = \frac{TP}{TP+FP+FN}$. For panoptic segmentation, we employ the mean Segmentation Quality (mSQ), mean Recognition Quality (mRQ) and mean Panoptic Quality (mPQ) \cite{Kirillov_2019_CVPR}. The mSQ measures the closeness of the predicted segments with their ground truths, mRQ is equivalent to the F1 score and the mPQ combines RQ and SQ: PQ = SQ$\times$RQ at a per-class level.
\paragraph{Application to image classification and object detection}
\textbf{Datasets}
We utilize the \textbf{Office-31 \cite{office31}} and Office-Home \cite{officehome} datasets to validate our proposal on image classification. The \textbf{Office-31} dataset contains 4,6K images composed of 31 object classes under three domains. The Office-Home dataset consists of 15,5K images of 65 categories  from four domains. 
Additionally, we also employ the UDA setup of Cityscapes \cite{Cordts2016Cityscapes}  to the Foggy Cityscapes daset \cite{SDV18} for object detection. Foggy Cityscapes is a synthetic foggy dataset which simulates fog on real scenes. Each foggy image is rendered with a clear image and depth map from Cityscapes. Thus the annotations and data split in Foggy Cityscapes are inherited from Cityscapes.

\subsection{Merging of models trained with same dataset and architecture}
This section investigates the merging of models trained on the same dataset and architecture for semantic segmentation. As merging methods, we compare Isotropic \cite{ModelAverage92, pmlr-v162-wortsman22a}, Fisher \cite{NEURIPS2022_70c26937}, output-ensemble \cite{HASHEM1997599} and our proposed layer-wise merging with different UDA strategies based on convolutional architectures (Advent \cite{vu2019advent}, MinEnt \cite{vu2019advent}, FADA \cite{Wang2020}, MaxSquare \cite{Chen_2019_ICCV} and  AdaptSegNet \cite{8578878}) and transformer architectures (DAFormer \cite{hoyer2022daformer}, HRDA \cite{hoyer2022hrda}, MIC \cite{hoyer2023mic} and PiPa \cite{chen2022pipa}). 

\begin{figure}[t]
    \centering
    
    \includegraphics[trim=1 1 1 1,clip,width=.7\linewidth]{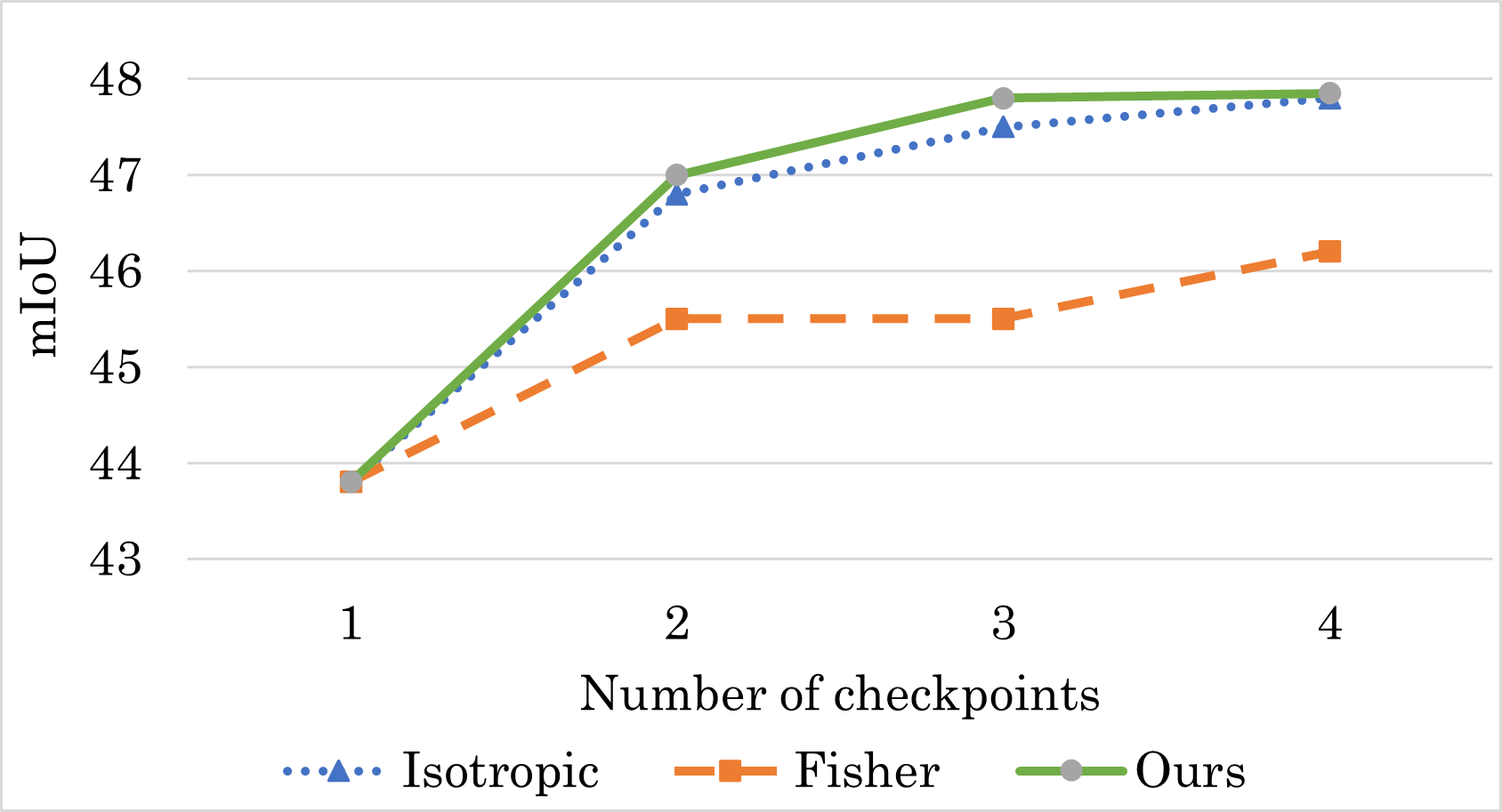}
    \caption{Performance comparison of model merging methods for different checkpoints of convolutional models on the GTA-to-Cityscapes setup. A single checkpoint represents the final performance (i.e., final checkpoint) of the UDA training strategy \cite{vu2019advent} with the DeepLabV2 architecture \cite{7913730}. Checkpoints greater than one consider merging the final checkpoint with an increasing number of checkpoints.}
    \label{fig:checkpoint_merge}
\end{figure}

\textbf{Same UDA strategy.}
Merging models from the same UDA strategy is conceptually similar to the update scheme of the teacher model in UDA \cite{hoyer2022daformer}. However, no previous study has quantified the actual advantages of merging checkpoints or analyzed different merging methods. Following \cite{ModelAverage92, li2023deep} our checkpoint selection process ensures a representative and homogeneous sampling of checkpoints. Specifically, we select checkpoints at evenly spaced intervals over the traing. As the anchor of our checkpoint merging strategy, we use the final checkpoint. Figure \ref{fig:checkpoint_merge} compares our results with Isotropic and Fisher alternatives for merging the training checkpoints, which are obtained every quarter of the total training iterations (i.e., 4 checkpoints in total). Then the merging is performed multiple times by incrementally employing the saved checkpoints. The model merging considers merging the final model weights (i.e., number of checkpoints equals to 1) and checkpoints obtained during training. As UDA strategies tend to present noisy training profiles, the merge of checkpoints aims at providing a more robust model which leverages all knowledge obtained during training. 

\begin{figure}[tp]
    \centering
     \begin{subfigure}[b]{0.25\linewidth}
     \caption{Color}
     \includegraphics[trim=0 200 0 0,clip,width=\linewidth]{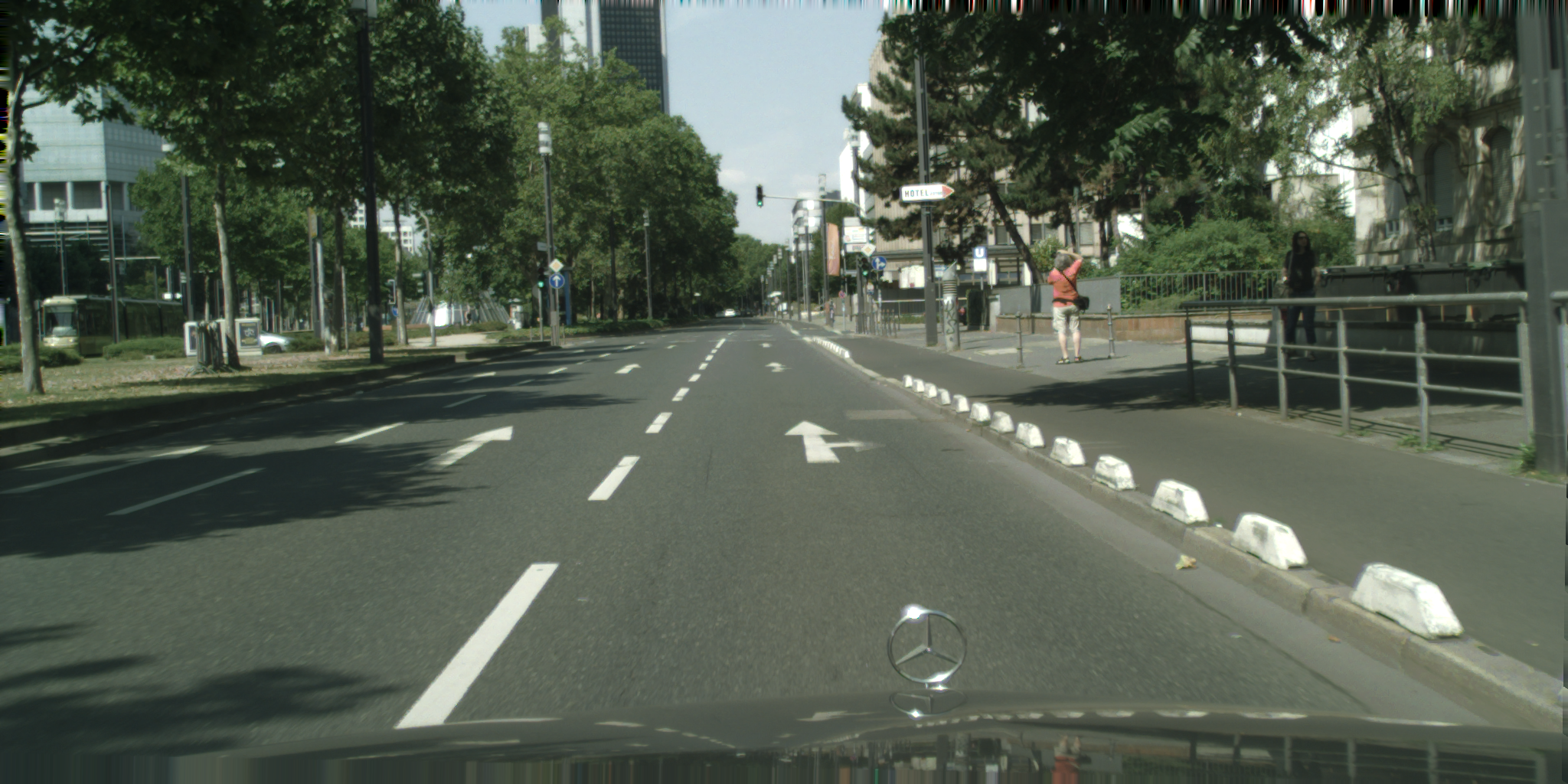}
     \end{subfigure}\hfill
    \begin{subfigure}[b]{0.25\linewidth}
    \caption{Ground truth}
     \includegraphics[trim=0 200 0 0,clip,width=\linewidth]{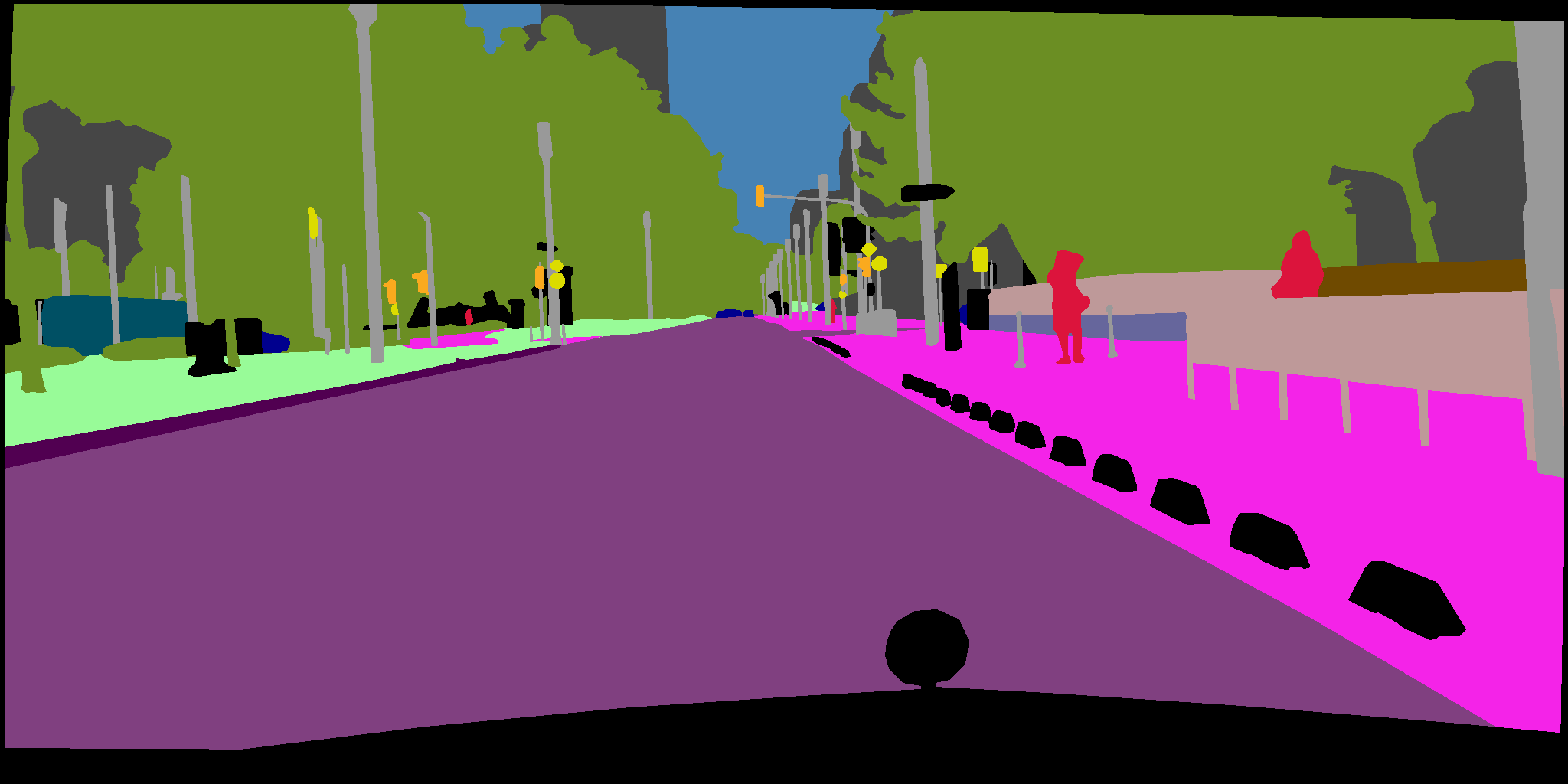}
     \end{subfigure}\hfill
     \begin{subfigure}[b]{0.25\linewidth}
     \caption{MIC \cite{hoyer2023mic}}
     \includegraphics[trim=0 200 0 0,clip,width=\linewidth]{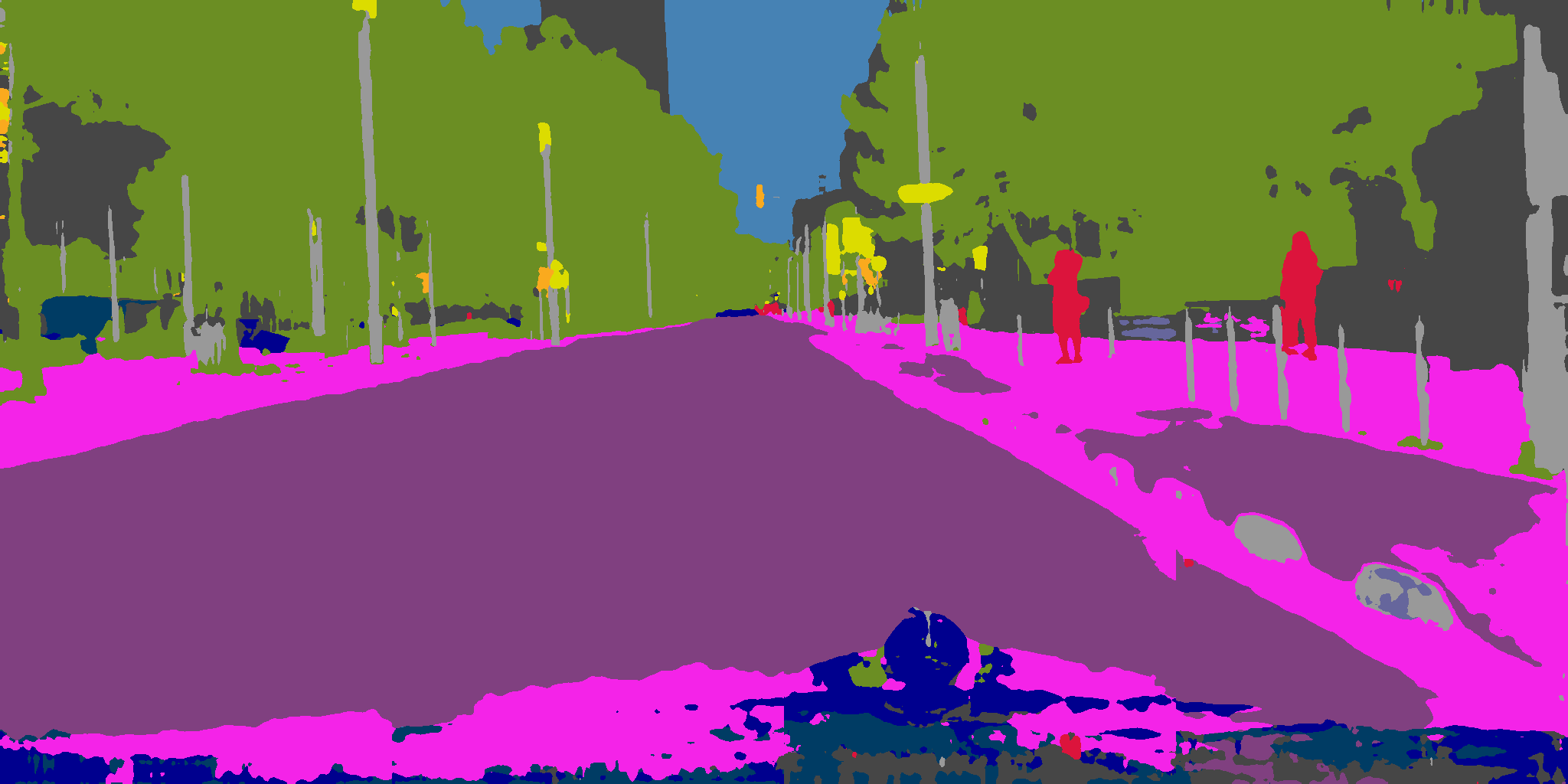}
     \end{subfigure}\hfill
     \begin{subfigure}[b]{0.25\linewidth}
     \caption{Ours}
     \includegraphics[trim=0 200 0 0,clip,width=\linewidth]{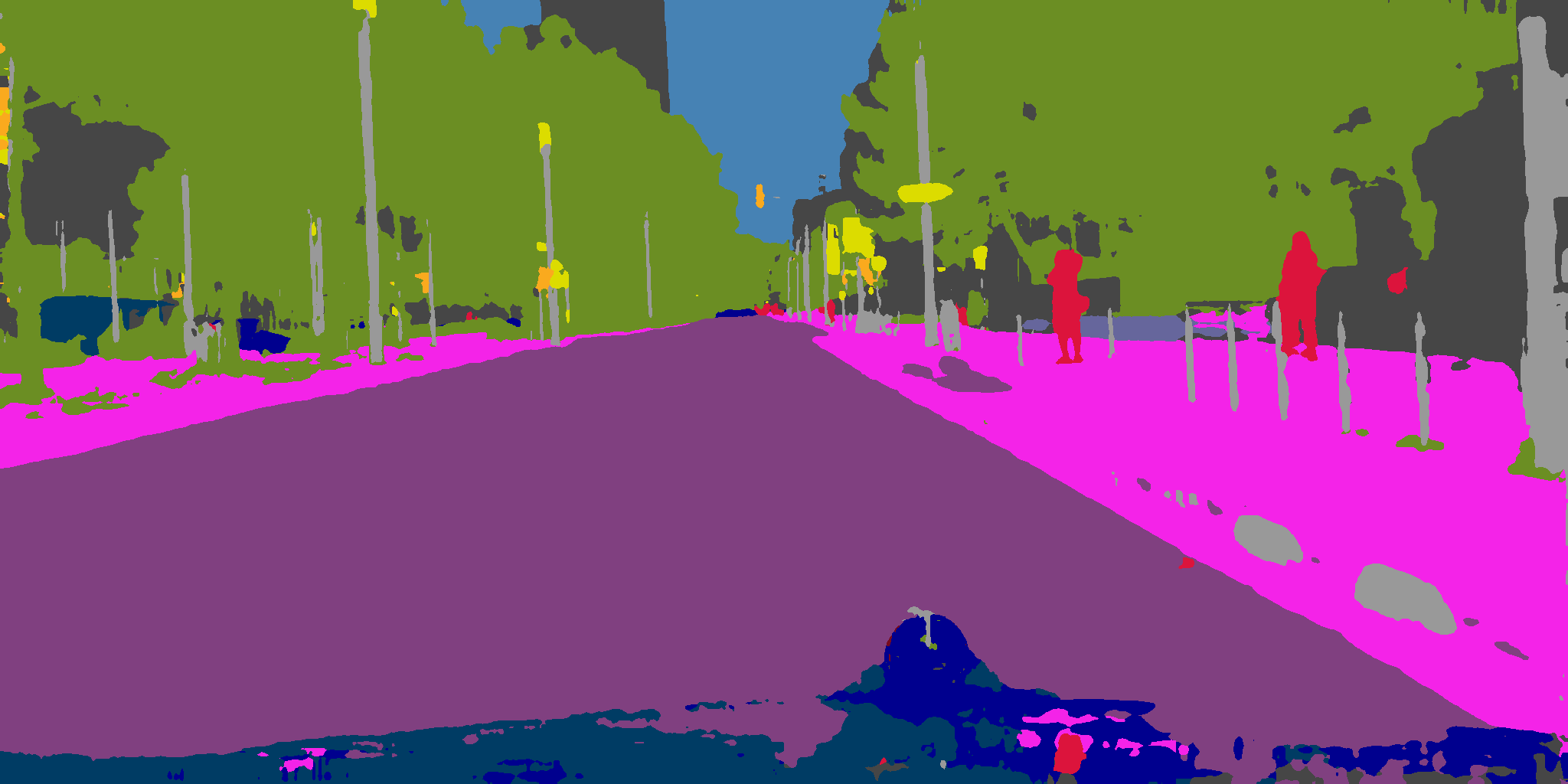}
     \end{subfigure}\hfill

     \begin{subfigure}[b]{0.25\linewidth}
     \includegraphics[trim=0 200 0 0,clip,width=\linewidth]{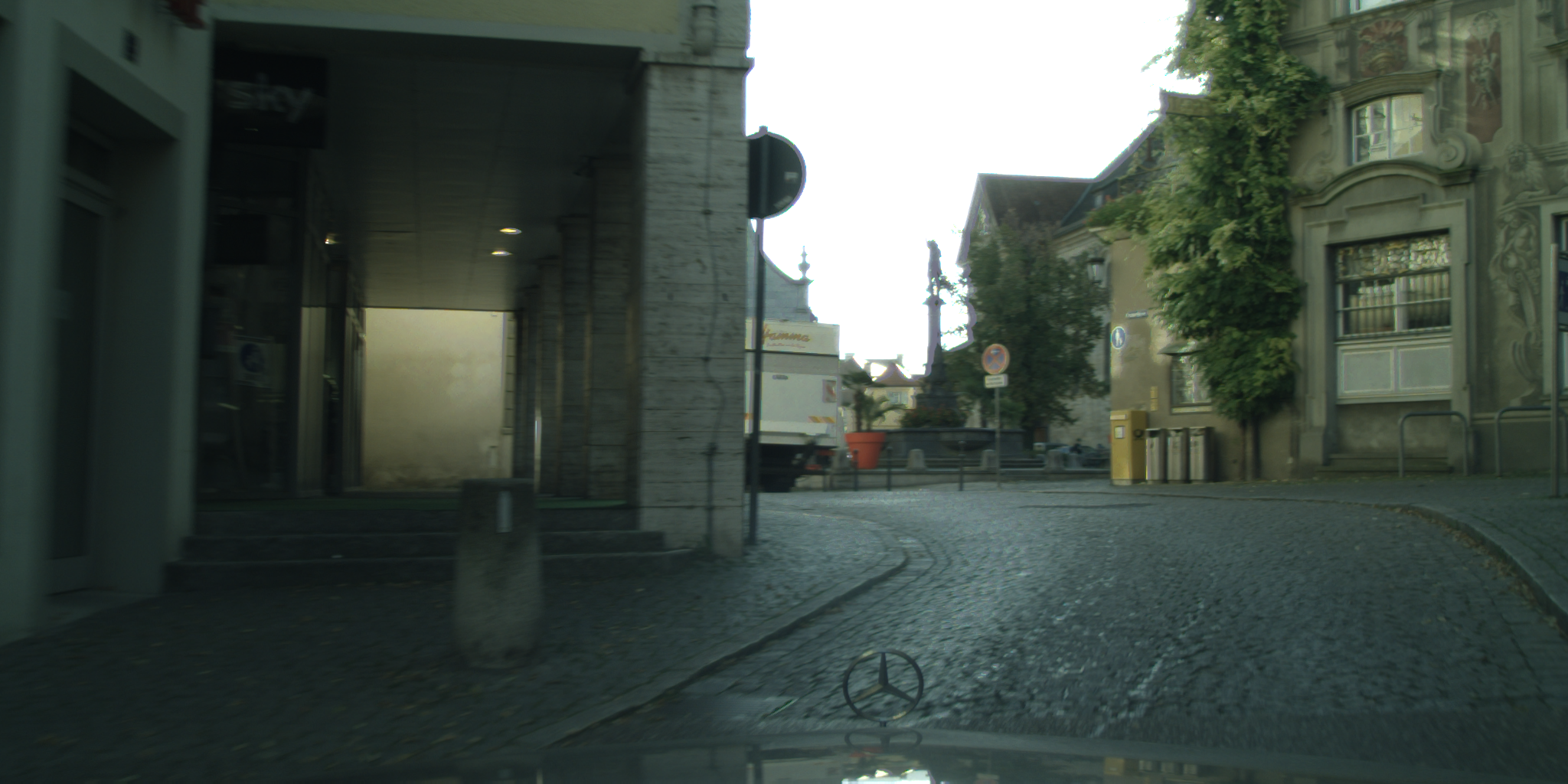}
     \end{subfigure}\hfill
    \begin{subfigure}[b]{0.25\linewidth}
     \includegraphics[trim=0 200 0 0,clip,width=\linewidth]{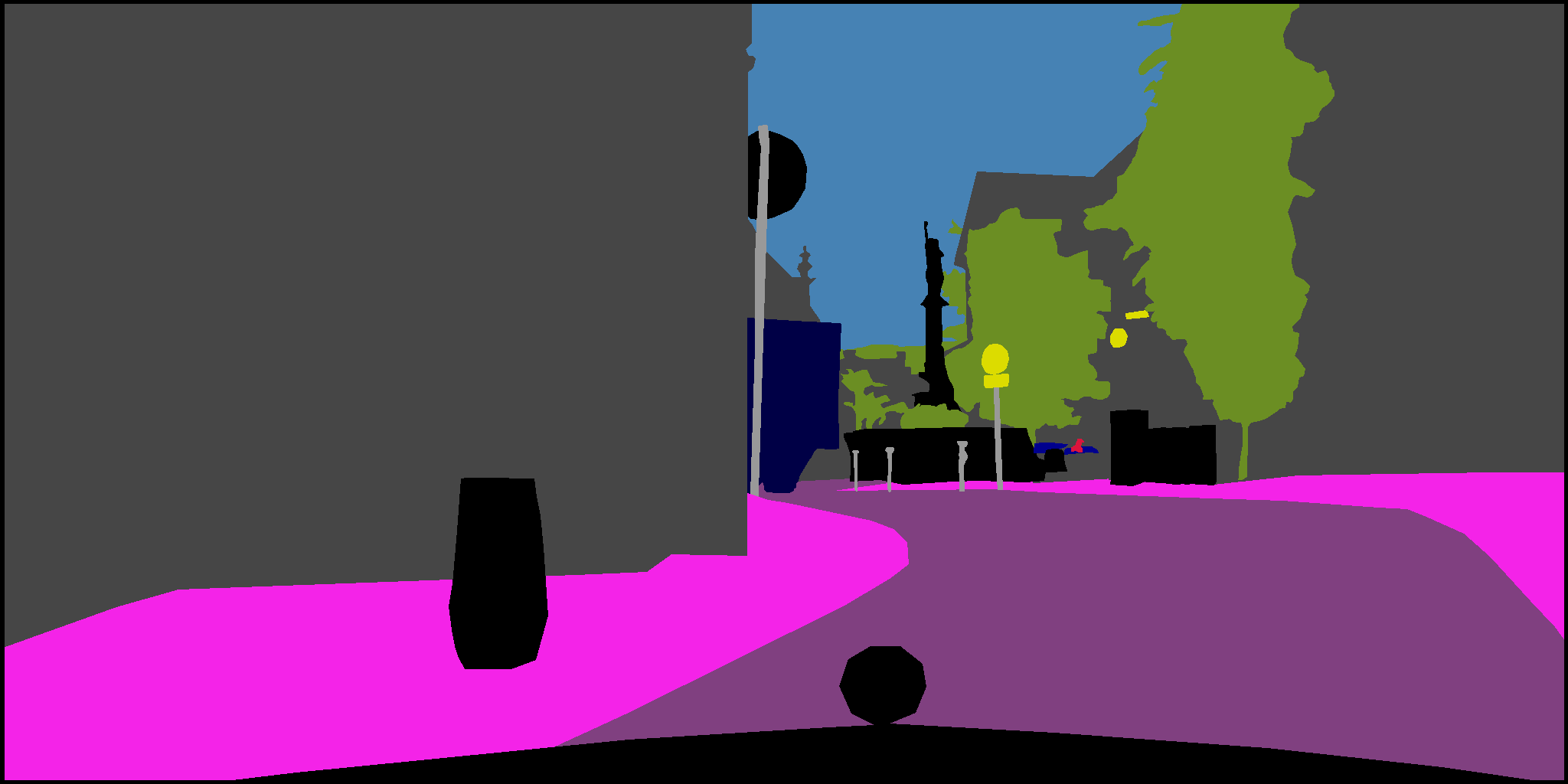}
     \end{subfigure}\hfill
     \begin{subfigure}[b]{0.25\linewidth}
     \includegraphics[trim=0 200 0 0,clip,width=\linewidth]{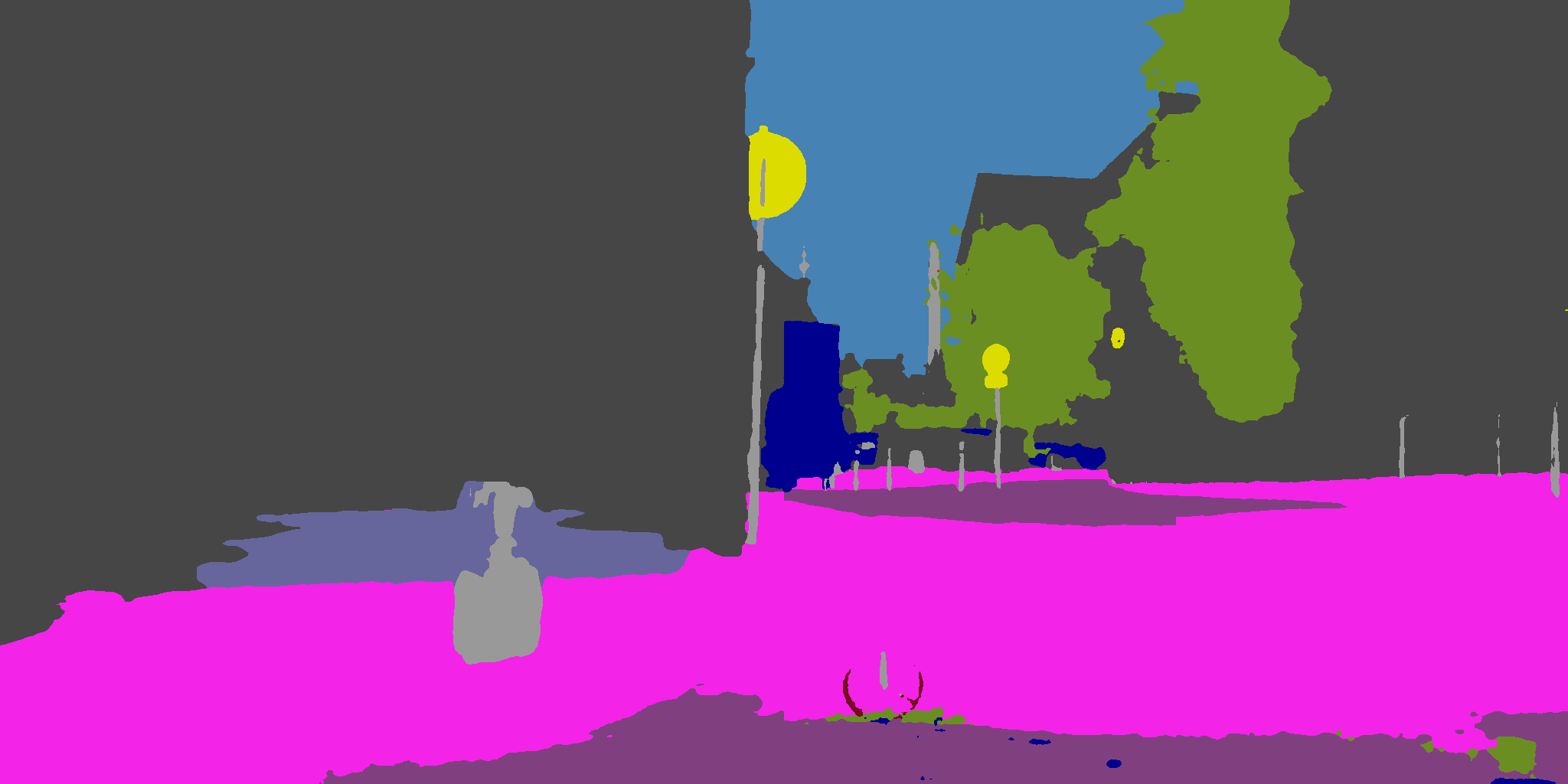}
     \end{subfigure}\hfill
     \begin{subfigure}[b]{0.25\linewidth}
     \includegraphics[trim=0 200 0 0,clip,width=\linewidth]{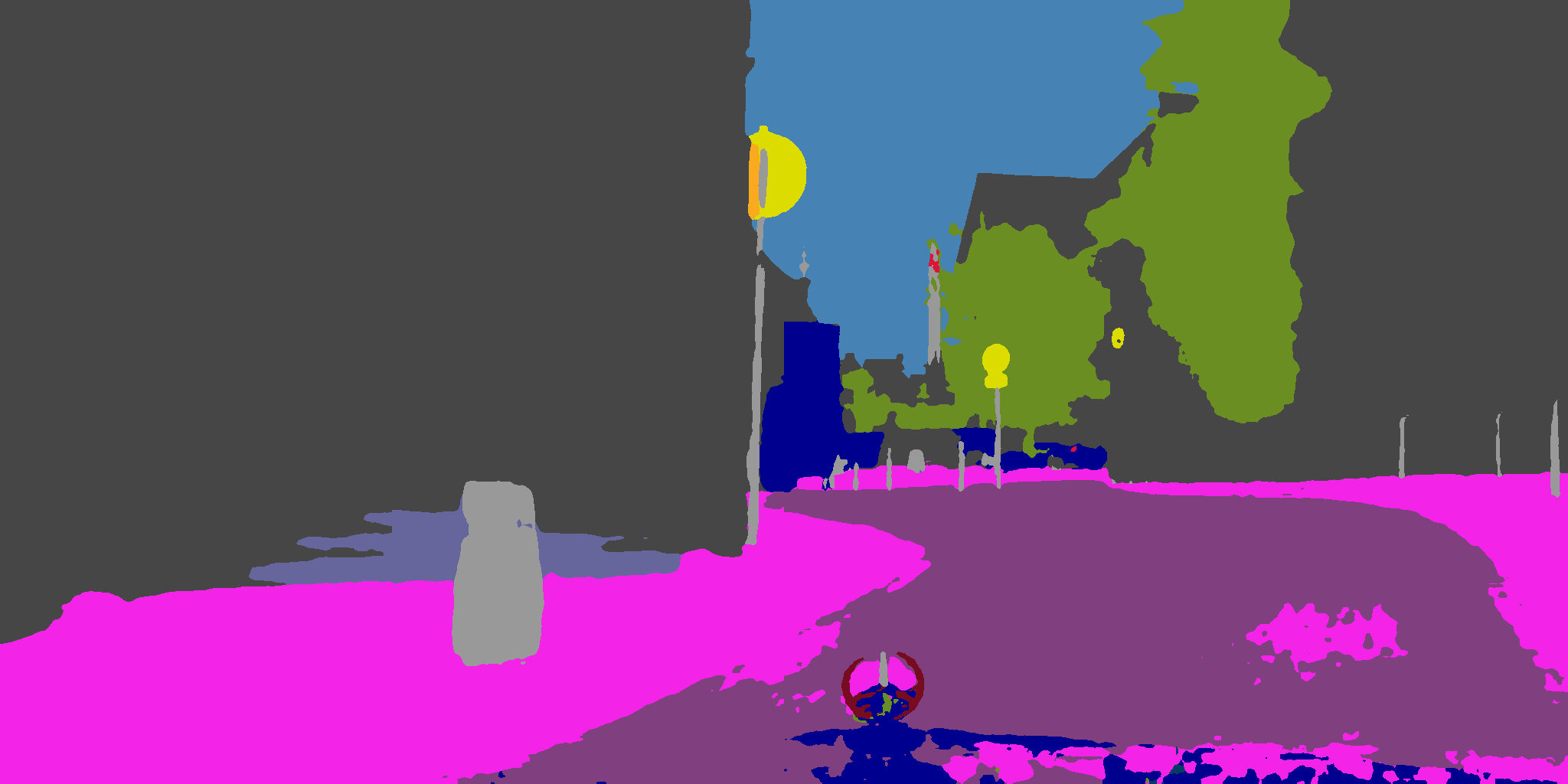}
     \end{subfigure}\hfill
   
    \begin{subfigure}[b]{0.25\linewidth}
     \includegraphics[trim=0 200 0 0,clip,width=\linewidth]{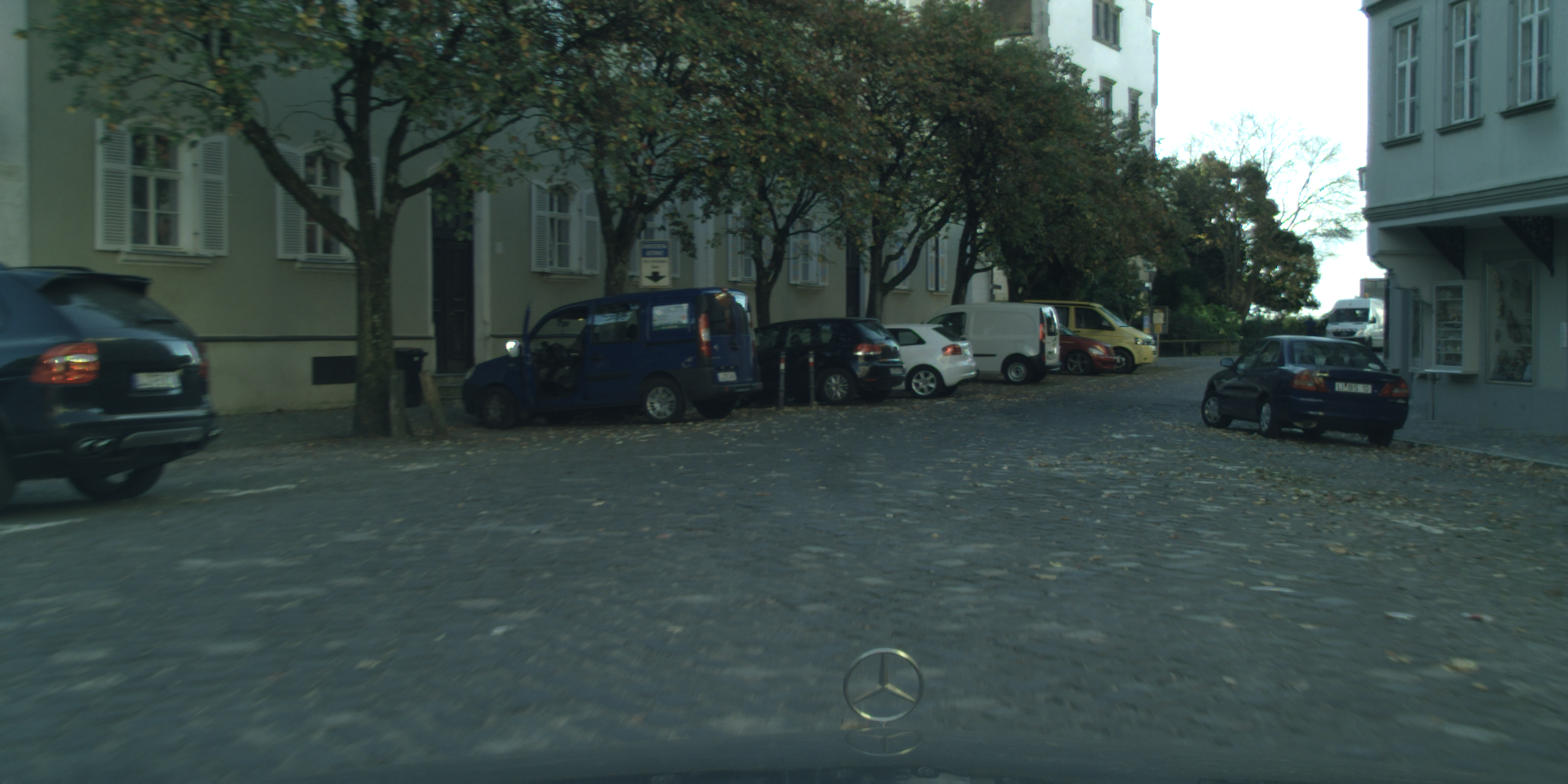}
     \end{subfigure}\hfill
    \begin{subfigure}[b]{0.25\linewidth}
     \includegraphics[trim=0 200 0 0,clip,width=\linewidth]{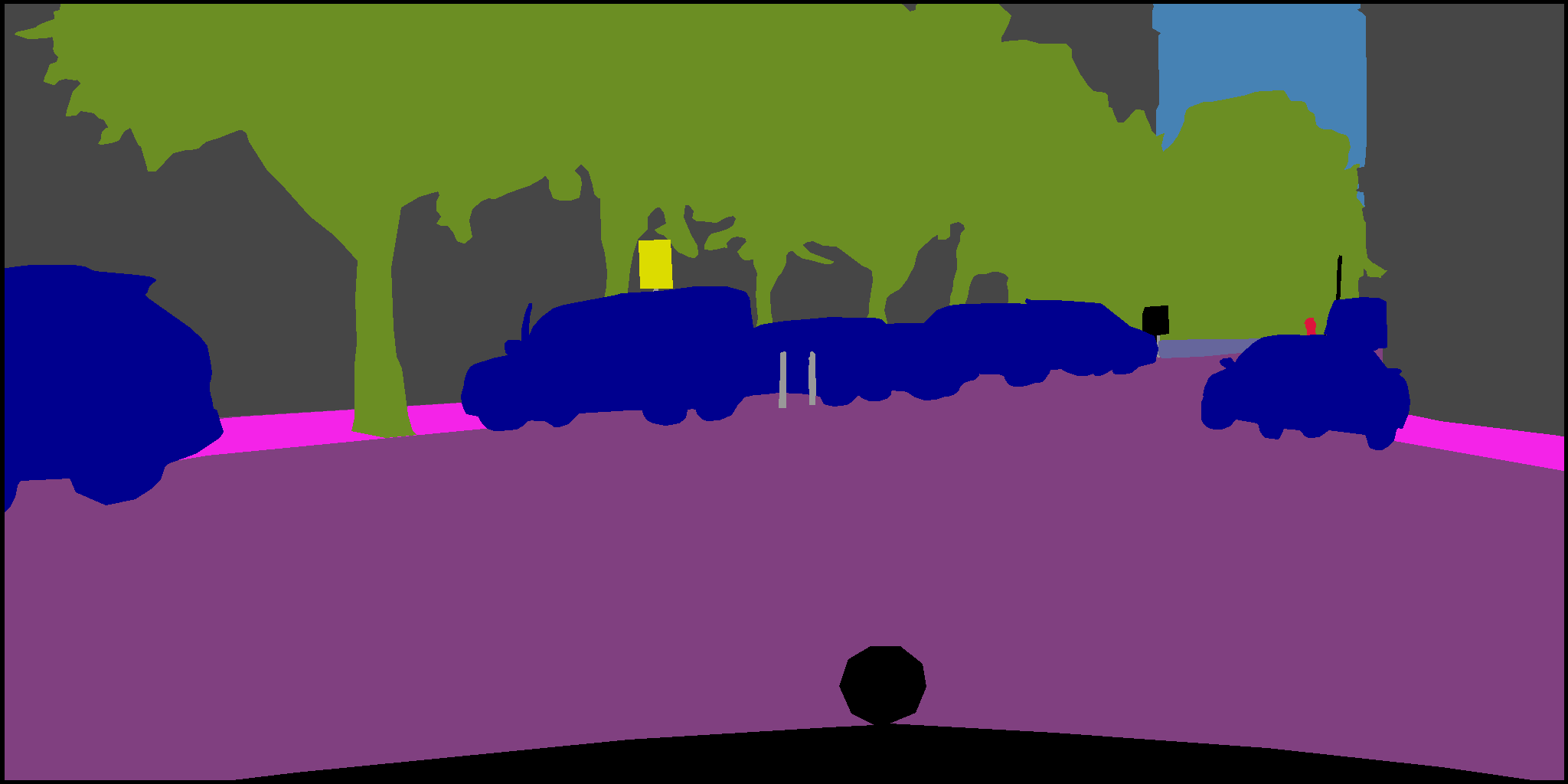}
     \end{subfigure}\hfill
     \begin{subfigure}[b]{0.25\linewidth}
     \includegraphics[trim=0 200 0 0,clip,width=\linewidth]{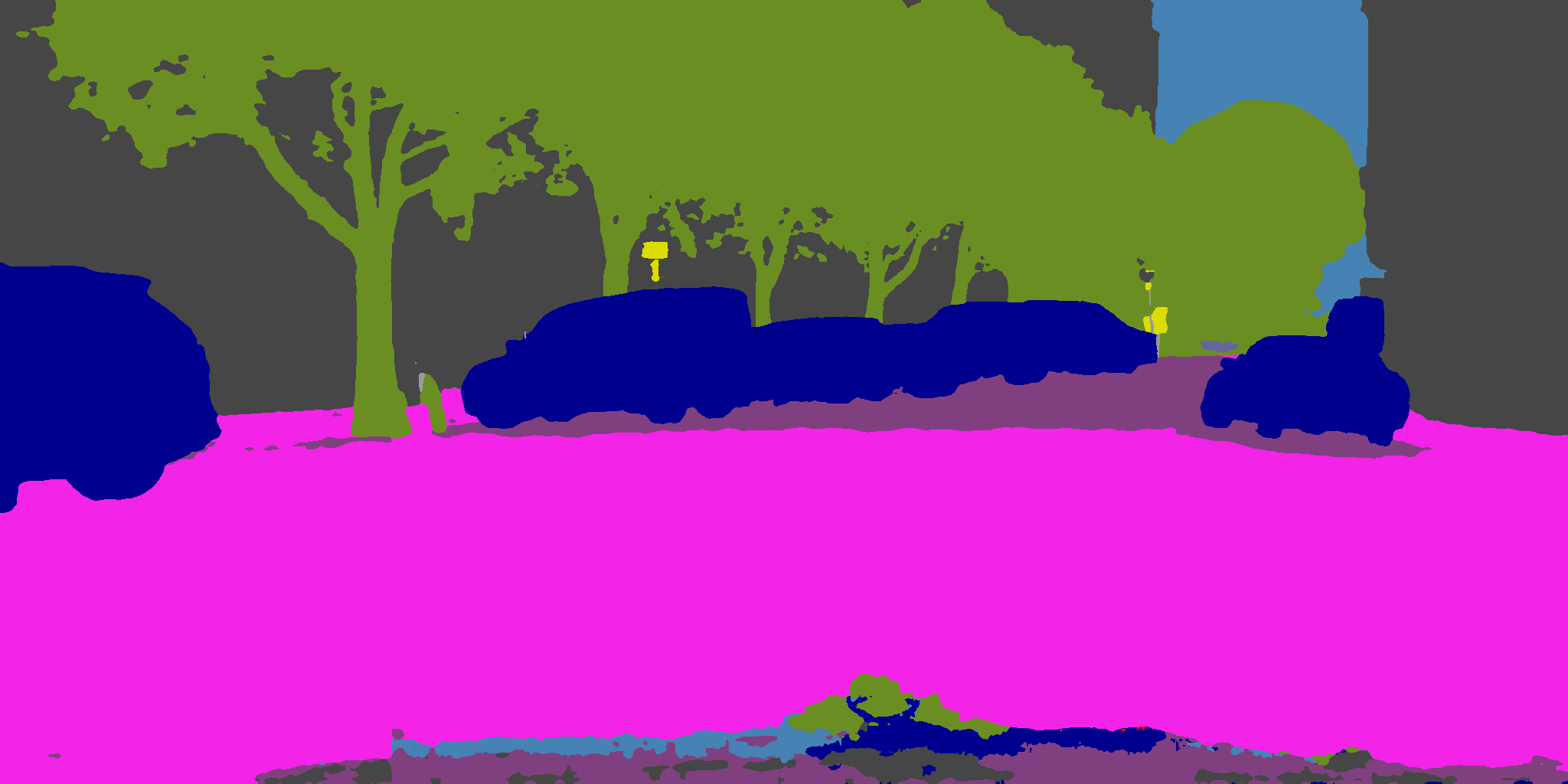}
     \end{subfigure}\hfill
     \begin{subfigure}[b]{0.25\linewidth}
     \includegraphics[trim=0 200 0 0,clip,width=\linewidth]{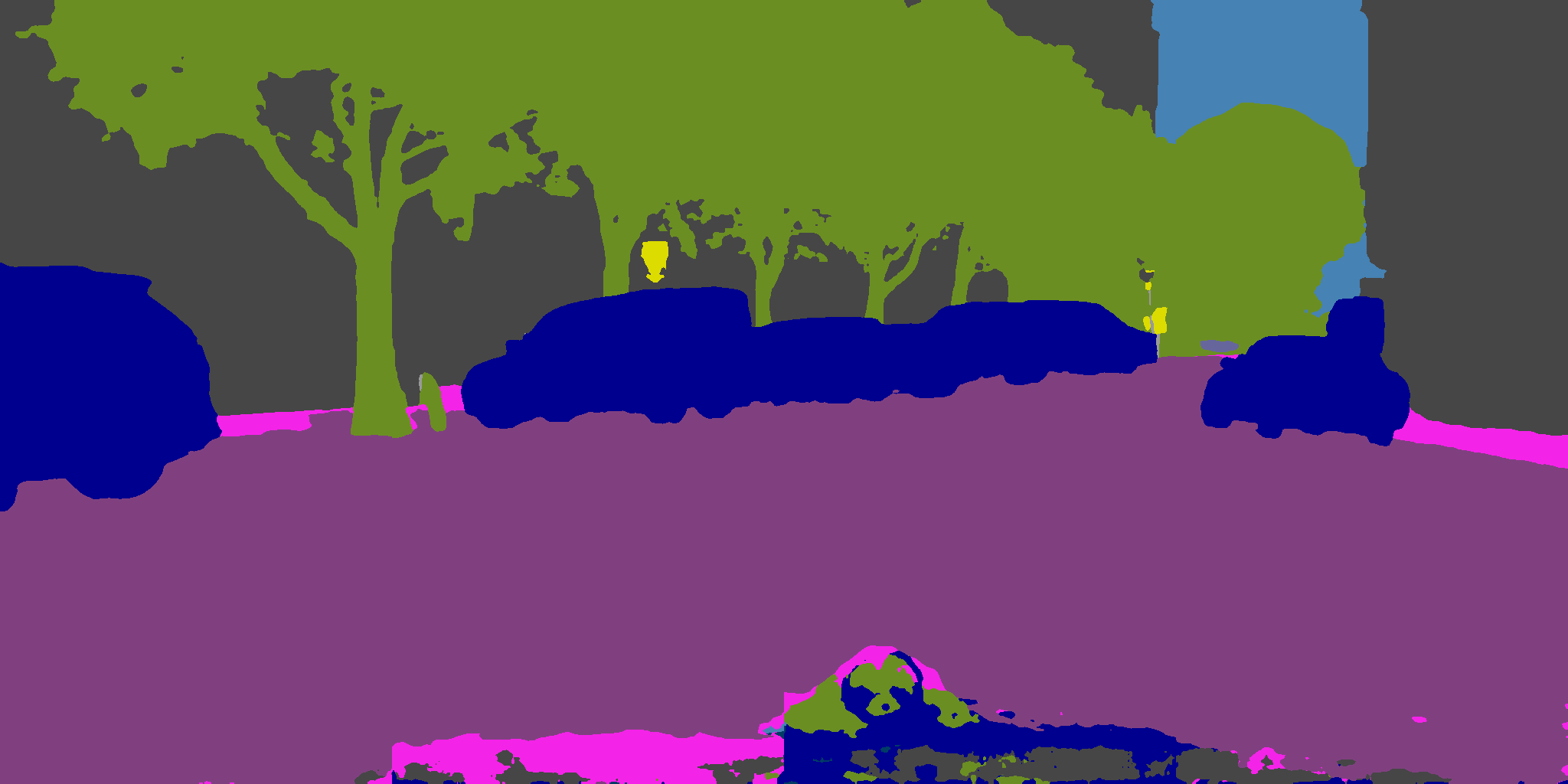}
     \end{subfigure}\hfill
     \caption{Qualitative comparison of state-of-the-art model MIC \cite{hoyer2023mic} on the GTA-to-Cityscapes UDA semantic segmentation setup for hard samples of coarse classes. Specifically, the sidewalk on a similar color than the road (first two rows), and a stone paved road (other four rows). Ours stands for the model merging of MIC checkpoints, note that our model is cost-free in terms of training time and inference time compared to the MIC model.  Each row presents the color image, the ground-truth labels, the segmentation result of MIC and our segmentation result (by columns).}
    \label{fig:segcomp}
\end{figure}

Figure \ref{fig:segcomp} showcases a qualitative comparison of our best segmentation results for merging checkpoints on the GTA-to-Cityscapes setup, against the top method MIC \cite{hoyer2023mic}. Notice how the state-of-the-art method struggles to present spatially robust outputs on the target domain. Here the shadow of the car or the texture of the terrain is wrongly segmented. Notably, the enhancements of our proposed merging of models enhances the consistency of the model by leveraging the model state on previous checkpoints. Therefore, improving results without recurring into computational overhead during training or testing, as the training and inference remain untouched.

\textbf{Different UDA strategy.}

Table \ref{tab:mergingSame} summarizes the performance comparison of the studied merging methods across five UDA training strategies for transformer and CNN-based models respectively. This experimental validation for merging models ranked from best to worst is employed in the literature to assess the trade-offs between performance enhancement and computational efficiency \cite{PIVA2023103745, pmlr-v162-wortsman22a}. Notably, all compared approaches decrease performance as more models are merged. However, our proposal distinguishes itself by consistently minimizing this decline, being the only one that significantly enhances the performance of individual models. This demonstrates the efficacy of our proposed merging strategy in maintaining robustness, even when integrating low-performance individual models.





\begin{table}[h!]
    \centering
    \resizebox{\linewidth}{!}{
    \setlength{\tabcolsep}{9pt}
    \begin{tabular}{c|c c c c c c}
        Number of &  \multicolumn{6}{c}{Transformer-based models}\\
         Models &Model mIoU& Isotropic& Fisher& Ensemble& Ours & $\Delta_{mIoU}$\\\toprule
         1& 75.9& \multicolumn{5}{c}{Anchor Model}\\
         2 & 75.6& 75.6&75.3&76.1&\textbf{76.2}& 1\% \\
         3&74.7&74.9&74.8&75.8&\textbf{76.3}& 2\% \\
         
         4&73.8& 74.8&74.3&75.2&\textbf{76.2}&2\%\\
         
         5&73.4& 74.2&73.9&74.7&\textbf{76.1}&3\%\\\midrule
         & \multicolumn{6}{c}{CNN-based models}\\\toprule
          1&46.9& \multicolumn{5}{c}{Anchor Model}\\
          2&45.2& 50.7& 47.3&51.0&\textbf{51.1} & 1\% \\
          3&43.8&48.7&46.8&51.1&\textbf{51.8}&6\% \\
          4&42.4& 48.2&47.3&48.7&\textbf{50.3} & 4\% \\
          5&42.3&47.7& 46.9& 48.1&\textbf{49.6}  & 4\% \\
         \bottomrule
    \end{tabular}
    }
    \caption{Performance comparison of model merging methods on the GTA-to-Cityscapes setup. These results consider including attractively each model sorted by their performance. Bold indicates best results for each number of models considered. $\Delta_{mIoU}$ compares our merging performance to the performance of Isotropic merging. CNN-based models employ DeepLab \cite{7913730} and Transformer-based models employ HRDA \cite{hoyer2022hrda}.}
    \label{tab:mergingSame}   
\end{table}
Table \ref{tab:mergingSame} also suggests that the merging of models trained with different UDA strategies provides better performances as compared to checkpoint merging. Specifically, the combination of CNN-based frameworks \cite{Wang2020,vu2019advent,8578878,Chen_2019_ICCV} outperforms by over 10\% the best-employed model \cite{Wang2020}. Meanwhile, transformer-based models mostly build-up from \cite{hoyer2022hrda} and present small drifts from one model to another, explaining the similar merging method's performance. However, as more methods are employed our merging is the only capable of preserving the knowledge from the anchor model \cite{hoyer2023mic}.

\textbf{Model merging improves performance of the anchor model regardless of the selected model}
In Table \ref{tab:mergingSameALL} we compare the performance of the merging across multiple merging schemes, UDA methods and different selection of anchor model. Notably, regardless of the employed methods, the merging improves the performance of the original methods. However, the combination of different types of UDA methods such as adversarial \cite{Wang2020} and entropy minimization \cite{Chen_2019_ICCV} showcase stronger performance than the combination of similar UDA methods such as two adversarial methods. These results suggest that our merging combines the knowledge of different training schemes to improve the final model performance.
\begin{table}[h!]
    \centering
    \setlength{\tabcolsep}{1pt}
    \resizebox{\linewidth}{!}{
    \setlength{\tabcolsep}{6pt}
    \begin{tabular}{ccccclllc}
        \multicolumn{5}{ c }{UDA method} & \multicolumn{4}{ c }{Merging method}\\\toprule
        Advent\cite{vu2019advent} & MinEnt\cite{vu2019advent} & FADA\cite{Wang2020} & MaxSquare\cite{Chen_2019_ICCV} & AdaptSegNet\cite{8578878} & Isotropic&Fisher&Ensemble&Ours \\\midrule
        $\checkmark$ & & & & &  43.8&43.8&43.8&43.8 \\ 
        & $\checkmark$ & & & &  42.3&42.3&42.3&42.3 \\ 
        & & $\checkmark$ & & &  46.9&46.9&46.9&46.9 \\ 
        & & & $\checkmark$ & & 45.2&45.2&45.2&45.2 \\ 
        & & & & $\checkmark$ &  42.4&42.4&42.4&42.4 \\ \midrule
        $\checkmark$ & $\checkmark$ & & & & 45.3 & 44.7& 45.4 & \textbf{45.4}\\ 
        $\checkmark$ & & $\checkmark$ & & & 48.0 & 42.3& 48.2& \textbf{48.2}  \\ 
        $\checkmark$ & & & $\checkmark$ & & 47.1 & 46.1& 47.3& \textbf{47.4}  \\ 
        $\checkmark$ & & & & $\checkmark$ & 44.6 & 44.0& \textbf{44.8}& 44.7\\ 
        $\checkmark$ & $\checkmark$ & $\checkmark$ & & & 47.3 & 46.9 & 47.9& \textbf{48.2} \\ 
        $\checkmark$ & $\checkmark$ & & $\checkmark$ & & 46.8 & 46.3 & 47.1& \textbf{47.2} \\ 
        $\checkmark$ & $\checkmark$ & & & $\checkmark$ & 45.0 & 44.7 & 46.5& \textbf{46.9} \\ 
        $\checkmark$ & $\checkmark$ & $\checkmark$ & $\checkmark$ & & 48.2 &47.3&48.7&\textbf{49.2}\\ 
        $\checkmark$ & $\checkmark$ & $\checkmark$ & & $\checkmark$ & 47.0 &45.3&47.7&\textbf{48.8}\\ 
        $\checkmark$ & $\checkmark$ & & $\checkmark$ & $\checkmark$ & 46.2 &45.2&46.3&\textbf{47.6}\\ 
        $\checkmark$ & $\checkmark$ & $\checkmark$ & $\checkmark$ & $\checkmark$ & 47.7 &46.9&48.1&\textbf{49.6}\\ 
        & $\checkmark$ & $\checkmark$ & & & 47.9 &46.1&48.0&\textbf{48.2}\\ 
        & $\checkmark$ & & $\checkmark$ & & 47.0 &47.1 & 47.5& \textbf{47.6} \\ 
        & $\checkmark$ & & & $\checkmark$ & 44.7 &44.3 & 45.1& \textbf{45.3}\\ 
        & $\checkmark$ & $\checkmark$ & $\checkmark$ & & 48.8&48.4&48.9&\textbf{48.9} \\ 
        & $\checkmark$ & $\checkmark$ & & $\checkmark$ & 47.0 &45.3&48.1&\textbf{49.2}\\ 
        & $\checkmark$ & & $\checkmark$ & $\checkmark$ & 46.3&44.8&46.9&\textbf{47.4} \\ 
        & $\checkmark$ & $\checkmark$ & $\checkmark$ & $\checkmark$ & 48.0 & 47.1& 48.8& \textbf{49.1} \\ 
        & & $\checkmark$ & $\checkmark$ & & 50.7 & 47.3&51.0&\textbf{51.1}\\ 
        & & $\checkmark$ & $\checkmark$ & $\checkmark$ & 48.7  & 46.8&49.1&\textbf{51.8}\\ 
        & & $\checkmark$ & & $\checkmark$ & 46.7 & 45.6&48.6&\textbf{50.4}\\ 
        \bottomrule
    \end{tabular}}
    \caption{Performance comparison of model merging methods employing models trained with different UDA methods but a common architecture. Results for the GTA-to-Cityscapes setup.}
    \label{tab:mergingSameALL}
\end{table}

\newpage
\textbf{Exploring merging parameters.}

\begin{figure}[tp]
    \centering
    \begin{subfigure}[b]{.33\linewidth}
        \includegraphics[width=\linewidth]{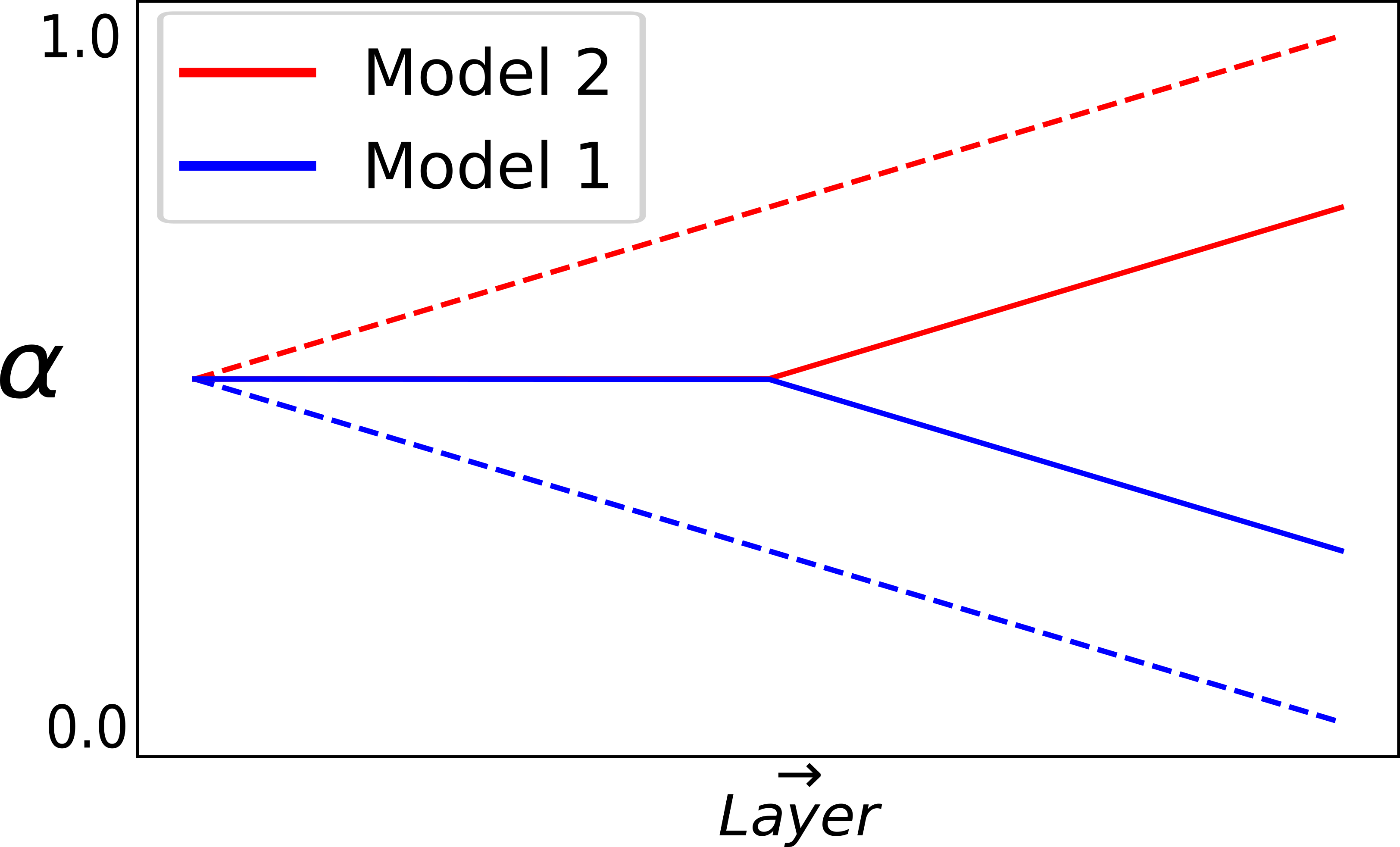}
        \caption{Visualization of the starting layer for merging weights.}
        \label{fig:explanation}
    \end{subfigure}\hfill
    \begin{subfigure}[b]{.3\linewidth}
    \includegraphics[width=\linewidth]{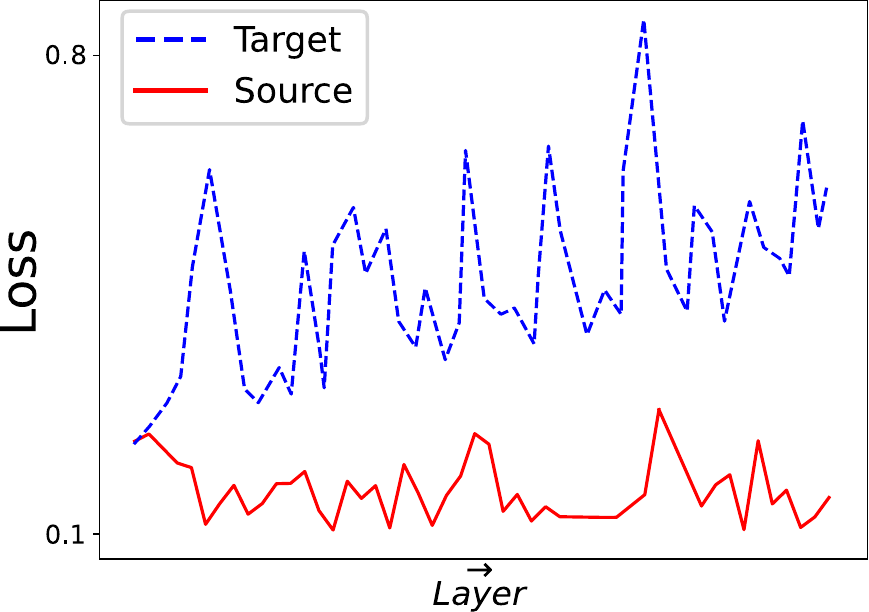}
        \caption{Performance comparison of the starting layer.}
        \label{fig:merging_loss}
    \end{subfigure}\hfill
    \begin{subfigure}[b]{.3\linewidth}
        \includegraphics[width=\linewidth]{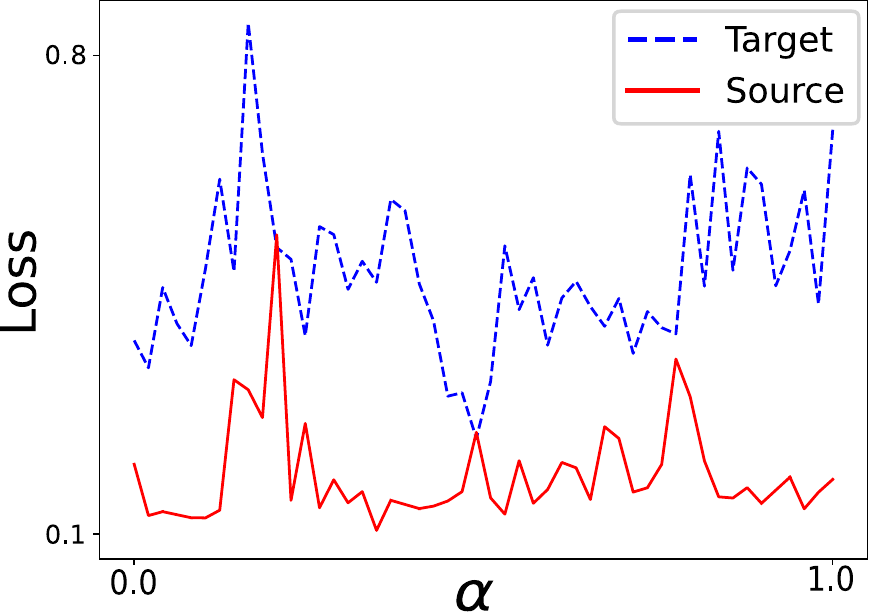}
        \caption{Starting layer weight performance comparison.}
        \label{fig:loss_alpha}
    \end{subfigure}
    \caption{In-depth analysis of Layer-wise merging of an adversarial \cite{Wang2020} and an entropy minimization \cite{vu2019advent} method on a DeepLabV2 architecture \cite{7913730}.}
\end{figure}

In developing our framework, we made two design decisions:
First, to determine the optimal starting layer and weight for assigning layer-wise weights, we conducted experiments visualized in Figure \ref{fig:explanation}. This figure contrasts the weight distribution between two models when initiating layer-wise merging at different layers. Additionally, Figure \ref{fig:merging_loss} demonstrates that introducing weight adjustments at later stages adversely affects the performance on the target dataset. Conversely, the performance on the source domain performance remains stable across various starting layers, indicating that our layer-wise merging technique predominantly enhances target dataset outcomes.

Second, the impact of varying the weights assigned to the initial layer is depicted in Figure \ref{fig:loss_alpha}, which shows the outcome of merging two models with differing initial weights, formalized as $\theta^{*(1)} = \alpha\theta_1^{(1)} + (1-\alpha)\theta_2^{(1)}$. Notably, we discovered that an $\alpha$ value of 0.5, representing an equal weighting approach, yields the highest performance on the target domain. This isotropic merging strategy, as compared in the main body of our work, demonstrates the effectiveness of balanced weight distribution in the initial layer.

Based on these experiments, we conclude that the source dataset's loss is an unreliable predictor of target dataset performance. This observation explains the ineffectiveness of Fisher merging  \cite{NEURIPS2022_70c26937} in merging UDA models.

\subsection{Merging of models trained with different segmentation heads.}
This section examines the merging of models trained for semantic segmentation with different classification heads, sharing a common backbone. This head disparity could be due to the varying number of classes in the trained datasets, as it changes the last classification layer, or because the classification head architecture is different. Table \ref{tab:comparison_model_merging_heads} compiles the experiments performed.

\textbf{Same head different number of classes.}
Table \ref{subtab1:model_merging_sameHeadDifferentData} presents the merging of GTA and Synthia datasets on the HRDA architecture, which have different numbers of classes (19 and 16 specifically). As previous solutions are not suited for merging models from different datasets, our layer-wise merging is the only solution which not only yields a segmentation model but a better one than each of the individual models.


\textbf{Different head architectures.}
Table \ref{subtab2:model_merging_differentHeadSameData} considers merging models with the same backbone but different segmentation heads. It demonstrates the merging of DAFormer and HRDA architectures, where only shared parameters are considered for merging and architecture specific layers remain untouched, following the Fisher protocol \cite{NEURIPS2022_70c26937}.  Our method is the only capable to improve performance, with significant different with alternatives of up to +70.2 mIoU. This outcome is to be expected as alternatives are not designed to merge distant models in the parameter space, especially those with varying classification heads.
\begin{table}[tp]
\centering
\begin{subtable}{.48\linewidth}
\centering
\resizebox{\linewidth}{!}{
\begin{tabular}{c|c|c|c}
\multicolumn{2}{c|}{Method-Dataset} & mIoU & $\Delta_{mIoU}$ \\
\hline
Datasets & GTAV & 73.8 & 0\\
Employed & Synthia & 66.1 & -7.7\\
\hline
            & Isotropic & 9.7 & -64.1\\
Merging     & Fisher    & 12.2 & -61.6\\
Models      & Ensemble  & 64.3 & -9.5\\
            & Ours      & \textbf{75.7} & \textbf{1.9}\\
\hline
\end{tabular}}
\caption{Different datasets - same head (HRDA).}
\label{subtab1:model_merging_sameHeadDifferentData}
\end{subtable}%
\hfill
\begin{subtable}{.48\linewidth}
\centering
\resizebox{\linewidth}{!}{
\begin{tabular}{c|c|c|c}
\multicolumn{2}{c|}{Method} & mIoU & $\Delta_{mIoU}$ \\
\hline
Individual & DAF & 68.3 & -7.6\\
Models & HRDA & 75.9 & 0\\
\hline
        & Isotropic & 6.0 & -69.9\\
Merging & Fisher    & 7.3 & -68.6\\
Models  & Ensemble  & 68.6 & -7.6\\
        & Ours      & \textbf{76.2} &\textbf{0.3} \\
\hline
\end{tabular}}
\caption{Different head - same dataset (GTAV).}
\label{subtab2:model_merging_differentHeadSameData}
\end{subtable}
\caption{Comparison for model merging methods with different heads due to the source datasets (GTA\cite{Richter_2016_ECCV} and Synthia \cite{Ros2016}) or head architectures (DAFormer\cite{hoyer2022daformer}, HRDA\cite{hoyer2022hrda}). Results on the Cityscapes validation set. Bold indicates best results. $\Delta_{mIoU}$ computed with respect to the anchor model (indicated by $\Delta_{mIoU}=0$).}
\label{tab:comparison_model_merging_heads}
\end{table}

\subsection{Potential benefits of model merging on different architectures}
Here, we present two strong benefits of merging models from different architectures: merging high-performing architectures into less accurate but faster architectures and transferring the knowledge of an easier task to a harder one.

\textbf{Merging high-performing architectures into less accurate but faster architectures}
Table \ref{tab:HRDAintoDAF} demonstrates the merging of HRDA, a high-performing model, into DAFormer, a shallower architecture. Our layer-wise merging achieves a mIoU of 73.0 while maintaining the same inference speed, as no additional parameters are included. This showcases the potential of our method in combining the strengths of different architectures.
\begin{table}[]
    \centering
    \setlength\tabcolsep{15pt}
    \resizebox{\linewidth}{!}{
    \begin{tabular}{l c c c c c}
        \multicolumn{2}{ c }{Backbone}& \multicolumn{2}{ c }{Head}&&Relative\\
        DAFormer & HRDA &DAFormer & HRDA & mIoU& inference \\\toprule
        $\checkmark$ & &$\checkmark$ & & 68.3 & 1 \\ 
        &$\checkmark$ & &$\checkmark$ & 75.9 & 3.25 \\ \hline
        
        $\checkmark$ &$\checkmark$ &$\checkmark$ && 73.0 & 1 \\
        $\checkmark$ &$\checkmark$ &&$\checkmark$ & 76.2 & 3.25 \\\bottomrule 
    \end{tabular}}
    \caption{Layer-wise merging of high performing models (HRDA \cite{hoyer2022hrda}) into a shallower architecture (DAFormer  \cite{hoyer2022daformer}). Relative inference speed measured with a Titan RTX GPU. Results are for the GTA-to-Cityscapes UDA semantic segmentation setup.}
    \label{tab:HRDAintoDAF}
\end{table}

\begin{figure}[h!]
    \begin{subfigure}[b]{0.25\linewidth}
    \caption{Color}
        \includegraphics[trim=60 1200 930 50,clip,width=\linewidth]{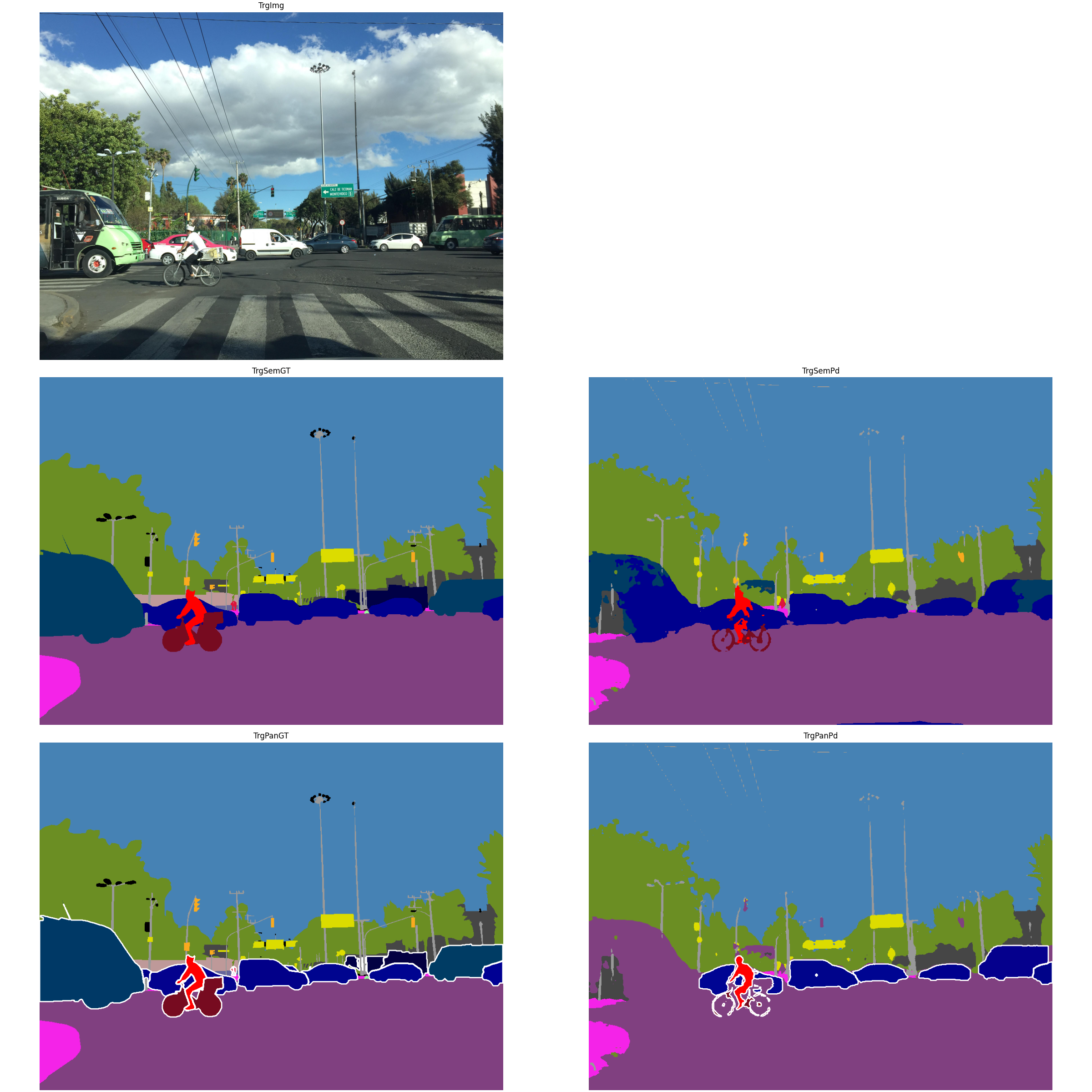}
    \end{subfigure}\hfill
    \begin{subfigure}[b]{0.25\linewidth}
    \caption{Ground truth}
        \includegraphics[trim=60 50 930 1200,clip,width=\linewidth]{imagenes/comparative/SOTA/panoptic/4VMNEgNxRwJWeApm8_Mfxg.png}
    \end{subfigure}\hfill
     \begin{subfigure}[b]{0.25\linewidth}
    \caption{EDAPS}
        \includegraphics[trim=930 50 60 1200,clip,width=\linewidth]{imagenes/comparative/SOTA/panoptic/4VMNEgNxRwJWeApm8_Mfxg.png}
    \end{subfigure}\hfill
    \begin{subfigure}[b]{0.25\linewidth}
    \caption{Ours}
        \includegraphics[trim=930 50 60 1200,clip,width=\linewidth]{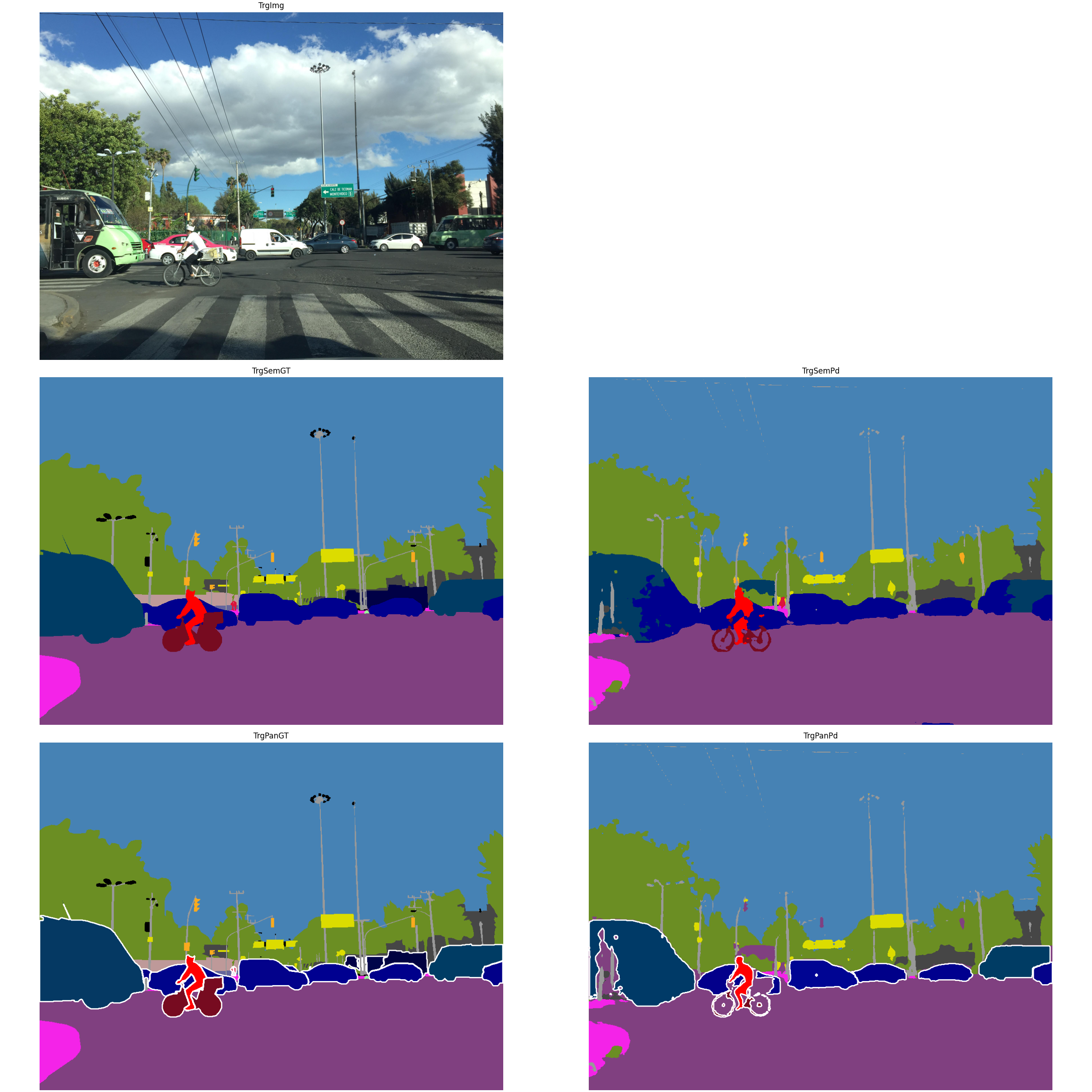}
    \end{subfigure}\\
    \begin{subfigure}[b]{0.25\linewidth}
        \includegraphics[trim=60 1200 930 50,clip,width=\linewidth]{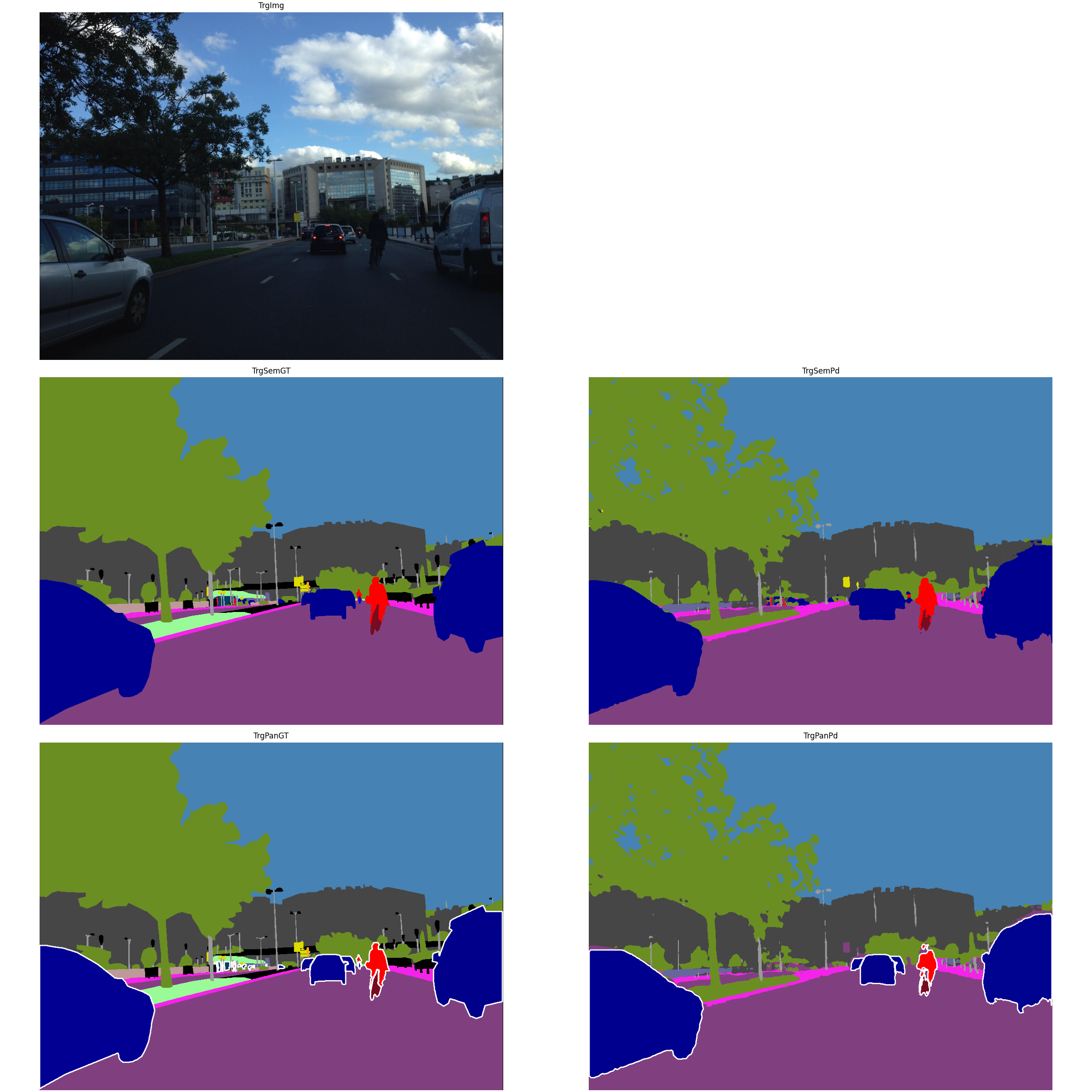}
    \end{subfigure}\hfill
    \begin{subfigure}[b]{0.25\linewidth}
        \includegraphics[trim=60 50 930 1200,clip,width=\linewidth]{imagenes/comparative/SOTA/panoptic/0VGACHmnzbjqiRs6BQ4ufA_structural.png}
    \end{subfigure}\hfill
     \begin{subfigure}[b]{0.25\linewidth}
        \includegraphics[trim=930 50 60 1200,clip,width=\linewidth]{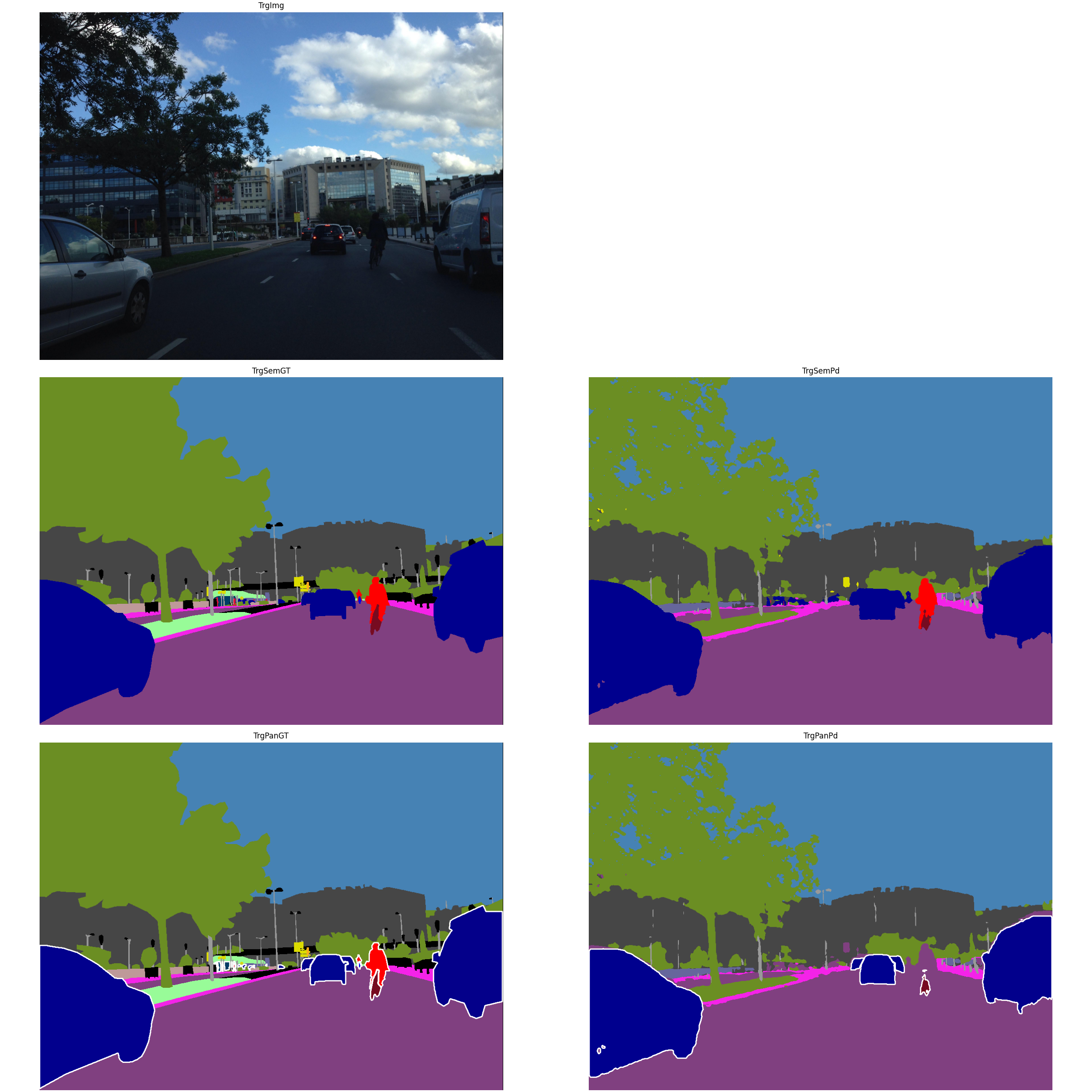}
    \end{subfigure}\hfill
    \begin{subfigure}[b]{0.25\linewidth}
        \includegraphics[trim=930 50 60 1200,clip,width=\linewidth]{imagenes/comparative/SOTA/panoptic/0VGACHmnzbjqiRs6BQ4ufA_structural.png}
    \end{subfigure}\\
    \begin{subfigure}[b]{0.25\linewidth}
        \includegraphics[trim=10 1300 870 80,clip,width=\linewidth]{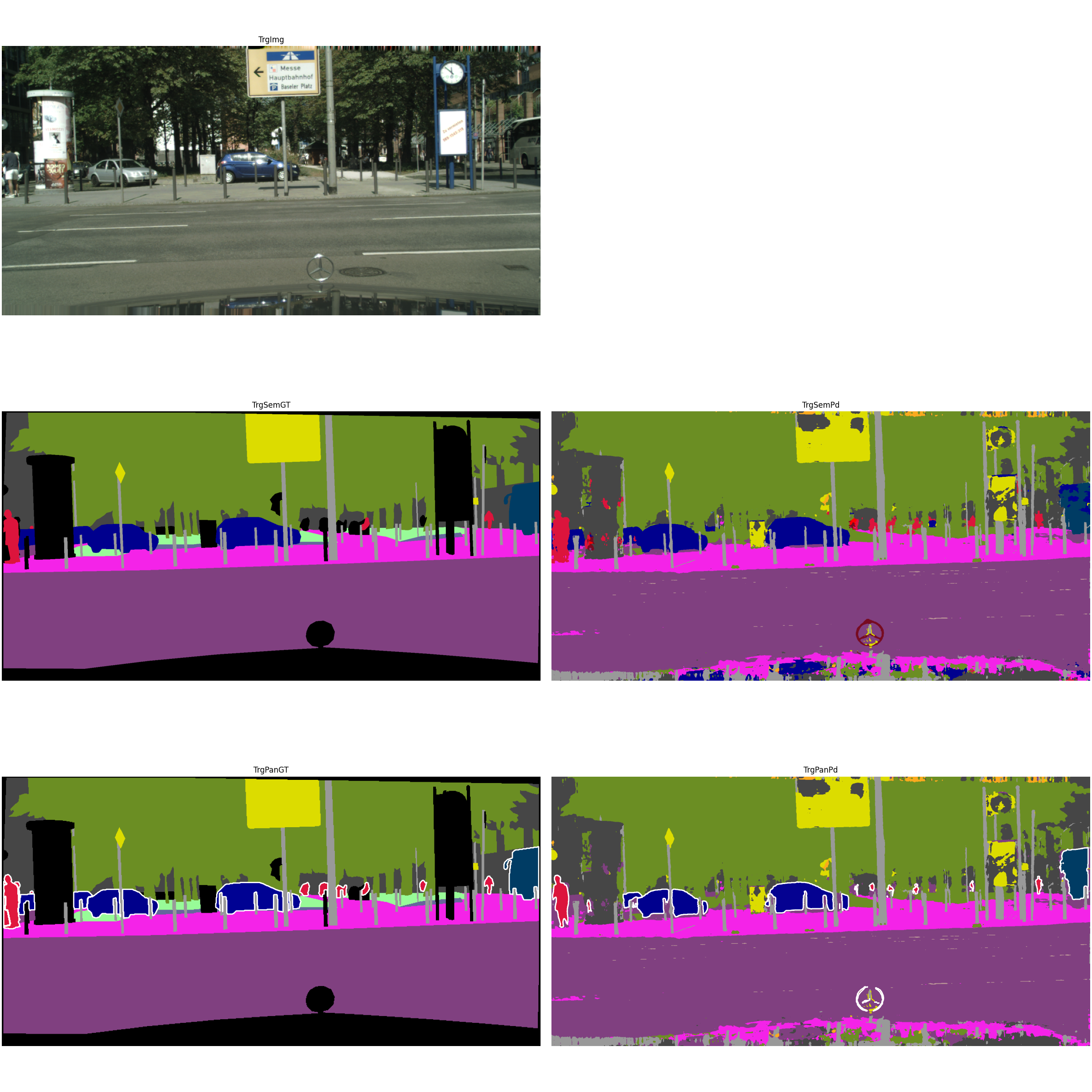}
    \end{subfigure}\hfill
    \begin{subfigure}[b]{0.25\linewidth}
        \includegraphics[trim=10 150 870 1230,clip,width=\linewidth]{imagenes/comparative/SOTA/panoptic/frankfurt_000000_011461_leftImg8bit.png}
    \end{subfigure}\hfill
     \begin{subfigure}[b]{0.25\linewidth}
        \includegraphics[trim=870 150 10 1230,clip,width=\linewidth]{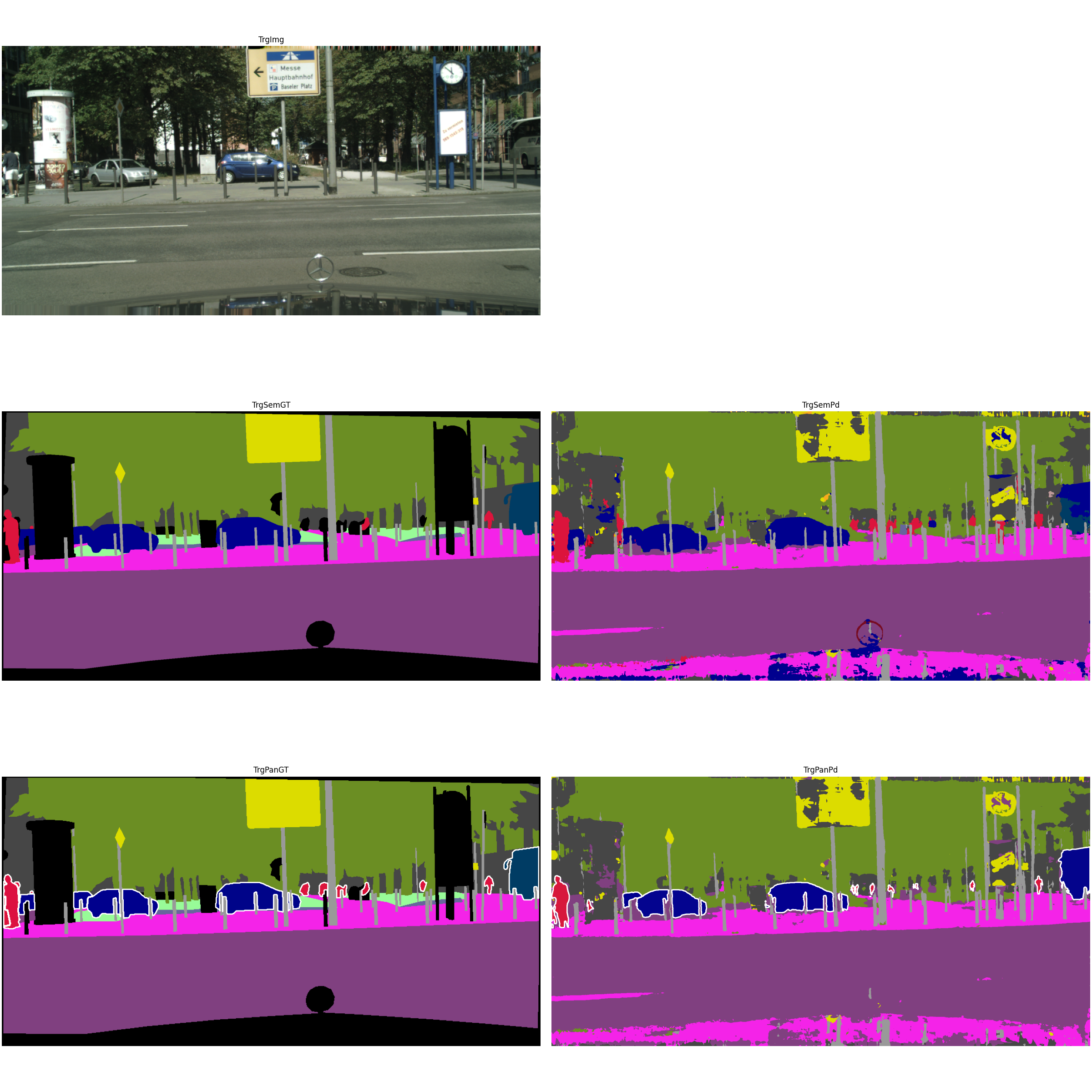}
    \end{subfigure}\hfill
    \begin{subfigure}[b]{0.25\linewidth}
        \includegraphics[trim=870 150 10 1230,clip,width=\linewidth]{imagenes/comparative/SOTA/panoptic/frankfurt_000000_011461_leftImg8bit.png}
    \end{subfigure}\\
    \begin{subfigure}[b]{0.25\linewidth}
        \includegraphics[trim=10 1300 870 80,clip,width=\linewidth]{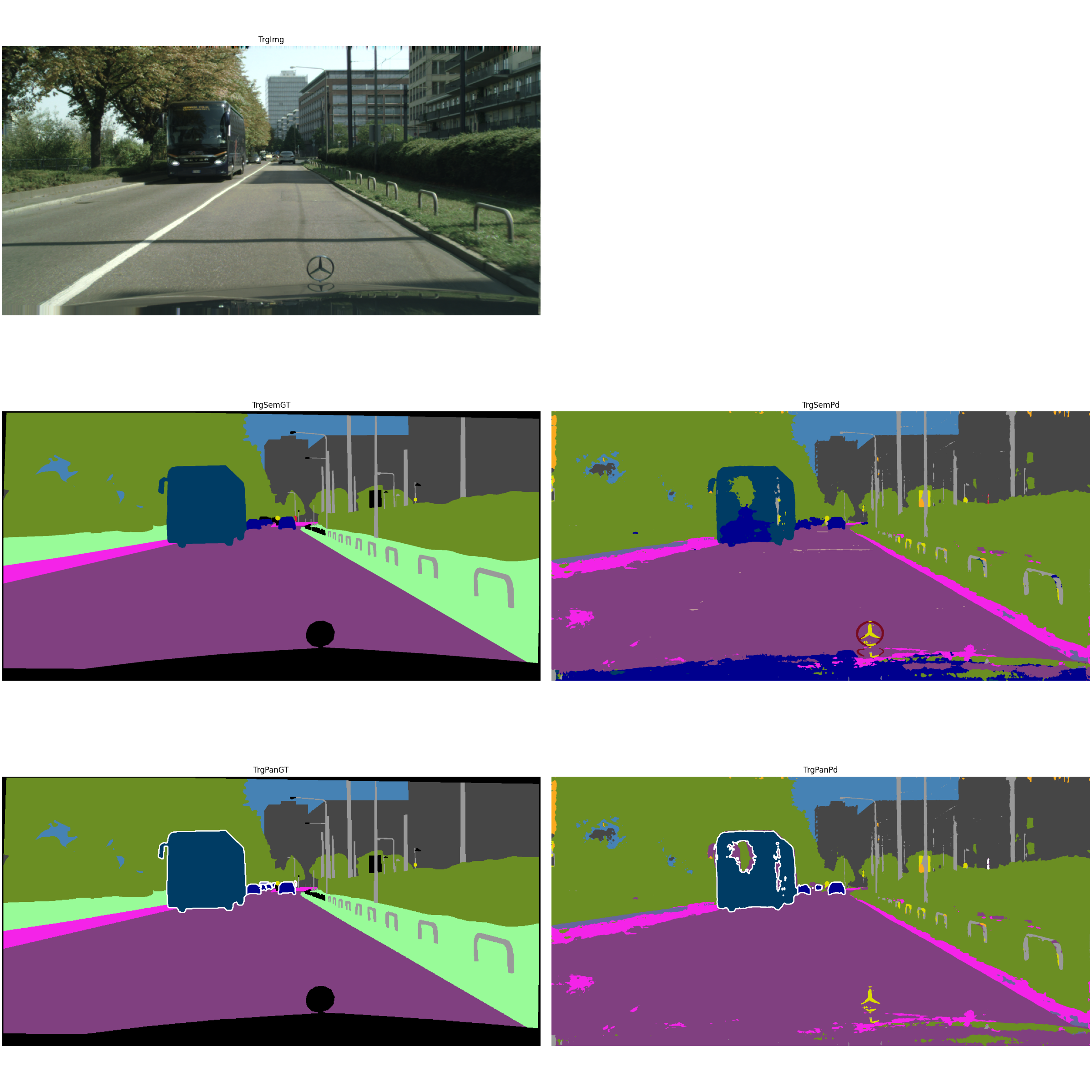}
    \end{subfigure}\hfill
    \begin{subfigure}[b]{0.25\linewidth}
        \includegraphics[trim=10 150 870 1230,clip,width=\linewidth]{imagenes/comparative/SOTA/panoptic/frankfurt_000000_007365_leftImg8bit.png}
    \end{subfigure}\hfill
     \begin{subfigure}[b]{0.25\linewidth}
        \includegraphics[trim=870 150 10 1230,clip,width=\linewidth]{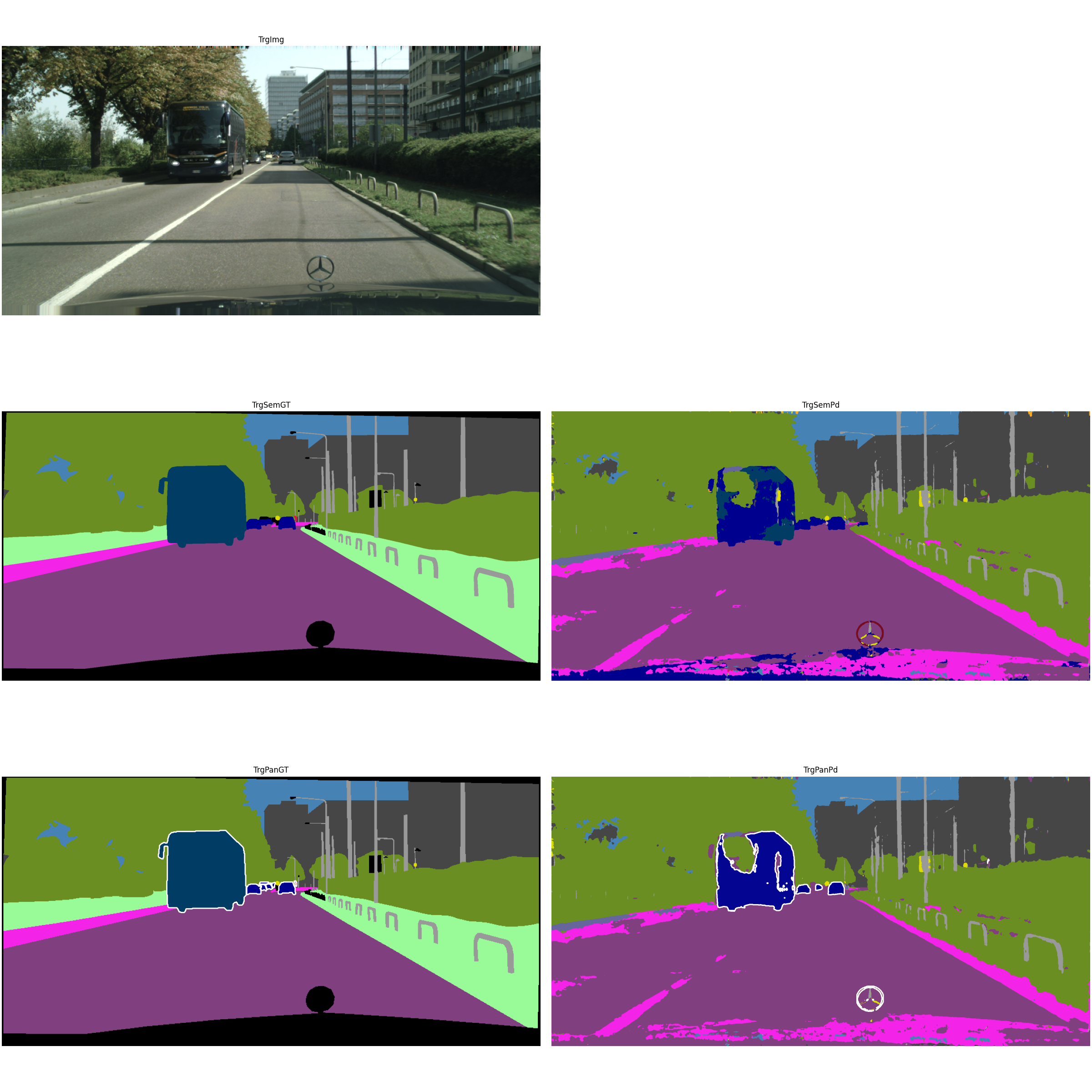}
    \end{subfigure}\hfill
    \begin{subfigure}[b]{0.25\linewidth}
        \includegraphics[trim=870 150 10 1230,clip,width=\linewidth]{imagenes/comparative/SOTA/panoptic/frankfurt_000000_007365_leftImg8bit.png}
    \end{subfigure}
    \caption{Qualitative comparison between state-of-the-art panoptic results of EDAPS \cite{edaps} and ours on the Synthia-to-Mapilliary (First two rows) and Synthia-toCityscapes (Bottom two rows) UDA setups for panoptic segmentation.}
    \label{fig:qualitative}
\end{figure}

\textbf{Merging knowledge between tasks}
Another possibility of model merging is the merging of models from semantic segmentation to panoptic segmentation. Specifically, Table \ref{tab:SemintoPan} presents the results of merging HRDA \cite{hoyer2022hrda} for semantic segmentation and EDAPS \cite{edaps} for panoptic segmentation. The resulting model improves across all metrics: +2.9 in mPQ, +3.0 in mRQ and +1.4 in mSQ the panoptic segmentation model\cite{edaps}.  Figure \ref{fig:qualitative} qualitatively showcases the segmentation capacities of our model compared to EDAPS. Specifically, our model is capable of detecting more instances than EDAPS (First and Second rows the \textit{bus} and \textit{rider} respectively) and is less confused between close looking classes (First, Third and Fourth rows, \textit{bus} and \textit{car}).  
\begin{table}[]
    \centering
    \setlength\tabcolsep{12pt}
    \resizebox{\linewidth}{!}{
    \begin{tabular}{l l l l l c l l c}
     \multicolumn{2}{ c }{Model} & \multicolumn{3}{ c }{Cityscapes} & \multicolumn{3}{ c }{ Mapilliary}\\\toprule
         EDAPS&HRDA&mSQ&mRQ&mPQ&mSQ&mRQ&mPQ\\\midrule
         $\checkmark$ & &72.7&53.6&41.2&71.7&46.1&36.6\\\hline
         $\checkmark$ &$\checkmark$ &74.3&56.6&44.1&74.3&48.5&38.5\\\bottomrule
    \end{tabular}}
    \caption{Layer-wise merging of models from different tasks: the semantic segmentation model HRDA \cite{hoyer2022hrda} into the panoptic segmentation one EDAPS \cite{edaps}. Results for the  Synthia-to-Cityscapes and Synthia-to-Mapilliary UDA panoptic segmentation setup.}
    \label{tab:SemintoPan}
\end{table}
    

\newpage

\subsection{Comparison with the state of the art}
Tables \ref{tab:SOTAsem} and \ref{tab:SOTApan} compare the per-class performance with state-of-the-art UDA methods for semantic and panoptic segmentation. Our merged models consistently present clear performance improvements on each benchmark. Specifically, we improve globally the state-of-the-art performance by +.4 mIoU on GTA2-to-Cityscapes, by +2.3 mIoU on Synthia-to-Cityscapes, by +2.9 mPQ on Synthia-to-Cityscapes, and by +1.9 mPQ on Synthia-to-Mapillary. Considering the class-wise performance in Tables \ref{tab:SOTAsem} and \ref{tab:SOTApan}, our models achieve consistent improvements for most classes when compared to the previous state-of-the-art methods across datasets and tasks. Compared to MIC \cite{hoyer2023mic} for semantic segmentation and EDAPS \cite{edaps} for panoptic segmentation, classes that have close lookalikes such as vehicles benefit the most. We argue that our layer-wise model merging enables the combination of the discriminative qualities of each merged model, thereby enhancing performance across visually related groups (e.g., up to +3\% mIoU and +11\% mPQ performance increase on vehicle type classes). 

\begin{table}[h!]
    \centering
    \resizebox{\textwidth}{!}{%
    \addtolength{\tabcolsep}{-5pt}  
    \begin{tabular}{l c c c c c c c c c c c c c c c c c c c c c }
         Method &\rotatebox[origin=c]{90}{\textit{road} } & \rotatebox[origin=c]{90}{\textit{sidewalk} } & \rotatebox[origin=c]{90}{\textit{building} } & \rotatebox[origin=c]{90}{\textit{wall}} &  \rotatebox[origin=c]{90}{\textit{fence}}& \rotatebox[origin=c]{90}{\textit{pole} } &\rotatebox[origin=c]{90}{\textit{light}}&\rotatebox[origin=c]{90}{\textit{sign} } & \rotatebox[origin=c]{90}{\textit{vegetation}} & \rotatebox[origin=c]{90}{\textit{terrain}}& \rotatebox[origin=c]{90}{\textit{sky}}& \rotatebox[origin=c]{90}{\textit{pedestrian}}& \rotatebox[origin=c]{90}{\textit{rider}} & \rotatebox[origin=c]{90}{\textit{car}}& \rotatebox[origin=c]{90}{\textit{truck}} & \rotatebox[origin=c]{90}{\textit{bus}}& \rotatebox[origin=c]{90}{\textit{train}} & \rotatebox[origin=c]{90}{\textit{motorcycle}} & \rotatebox[origin=c]{90}{\textit{bicycle}}& mean\tabularnewline\midrule
         &\multicolumn{19}{ c }{GTA-to-Cityscapes}&\\\midrule
         Wang\cite{expCons}& 96.5& 73.9&89.5&56.8&48.9&50.7&55.8&63.3&89.9&49.1&91.2&72.2&45.4&92.7&78.3&82.9&67.5&55.2&63.4&69.6\\
         HRDA\cite{hoyer2022hrda}&96.4&74.4&91.0&\underline{61.6}&51.5&57.1&63.9&69.3&91.3&48.4&94.2&79.0&52.9&93.9&84.1&85.7&75.9&63.9&67.5&73.8\\
         DIGA\cite{Shen_2023_CVPR}&97.0&78.6&91.3&60.8&\underline{56.7}&56.5&64.4&69.9&91.5&50.8&93.7&79.2&55.2&93.7&78.3&86.9&77.8&63.7&65.8&74.3\\
         
         MIC\cite{hoyer2023mic} & \underline{97.4} & \underline{80.1} & \underline{91.7} & 61.2 & \textbf{56.9} & \underline{59.7} &  \textbf{66.0} & \underline{71.3} & \underline{91.7} & \underline{51.4} & \textbf{94.3} & \underline{79.8} & \textbf{56.1} & \underline{94.6} & \underline{85.4} & \underline{90.3} & \underline{80.4} & \underline{64.5} & \textbf{68.5} & \underline{75.9} \\
         
        Ours & \textbf{97.6} & \textbf{81.3} & \textbf{91.8} & \textbf{63.0} & 56.3 & \textbf{60.7} & \underline{65.0} & \textbf{71.3} & \textbf{91.7} & \textbf{52.0} & \underline{94.0} & \textbf{80.1} & \underline{56.0} & \textbf{94.6} & \textbf{86.4} & \textbf{90.6} & \textbf{82.2} & \textbf{65.9} & \underline{68.4} & \textbf{76.3} \\ \hline

         &\multicolumn{19}{ c }{Synthia-to-Cityscapes}&\\\midrule
         Wang\cite{expCons}&83.7&42.9&87.4&39.8&7.5&50.7&55.7&53.5&85.9&&90.9&74.5&47.2&86.0&&60.2&&57.8&60.8&61.5\\
         HRDA\cite{hoyer2022hrda}&85.2&47.7&88.8&49.5&4.8&57.2&65.7&60.9&85.3&&92.9&79.4&52.8&89.0&&64.7&&63.9&\underline{64.9}&65.8\\
         DIGA\cite{Shen_2023_CVPR}&\underline{88.5}&\underline{49.9}&\underline{90.1}&\textbf{51.4}&6.6&55.3&64.8&62.7&\underline{88.2}&&93.5&78.6&51.8&89.5&&62.2&&61.0&\textbf{65.8}&66.2\\
         MIC\cite{hoyer2023mic}&84.3&45.6&90.1&48.8&\underline{9.2}&\underline{60.8}&\underline{66.8}&\underline{64.4}&87.4&&\underline{94.4}&\textbf{81.4}&\textbf{58.0}&\underline{89.7}&&\textbf{65.2}&&\underline{67.1}&64.1&\underline{67.5}\\
         Ours&\textbf{91.1}&\textbf{57.1}&\textbf{90.3}&\underline{50.4}&\textbf{9.2}&\textbf{61.3}&\textbf{67.0}&\textbf{65.7}&\textbf{89.4}&&\textbf{94.9}&\underline{80.6}&\underline{57.7}&\textbf{90.2}&&\underline{65.1}&&\textbf{67.4}&64.6&\textbf{69.8}\\\bottomrule
         \end{tabular}}
         \caption{Per-class performance comparison with state-of-the-art UDA methods on semantic  segmentation for the GTA-to-Cityscapes and Synthia-to-Cityscapes setups. Our proposal corresponds to the layer-wise model merging of MIC\cite{hoyer2023mic} and HRDA\cite{hoyer2022hrda}. Best results are indicated with bold while second best results are underlined.}
    \label{tab:SOTAsem}
     \vspace{-5mm}
\end{table}

\begin{table}[h!]
    \centering
    \resizebox{\textwidth}{!}{%
    \addtolength{\tabcolsep}{-5pt}  
    \begin{tabular}{l c c c c c c c c c c c c c c c c c c c c c }
         Method &\rotatebox[origin=c]{90}{\textit{road} } & \rotatebox[origin=c]{90}{\textit{sidewalk} } & \rotatebox[origin=c]{90}{\textit{building} } & \rotatebox[origin=c]{90}{\textit{wall}} &  \rotatebox[origin=c]{90}{\textit{fence}}& \rotatebox[origin=c]{90}{\textit{pole} } &\rotatebox[origin=c]{90}{\textit{light}}&\rotatebox[origin=c]{90}{\textit{sign} } & \rotatebox[origin=c]{90}{\textit{vegetation}}& \rotatebox[origin=c]{90}{\textit{sky}}& \rotatebox[origin=c]{90}{\textit{pedestrian}}& \rotatebox[origin=c]{90}{\textit{rider}} & \rotatebox[origin=c]{90}{\textit{car}}& \rotatebox[origin=c]{90}{\textit{bus}} & \rotatebox[origin=c]{90}{\textit{motorcycle}} & \rotatebox[origin=c]{90}{\textit{bicycle}}&mSQ&mRQ& mPQ\tabularnewline\midrule
         &\multicolumn{19}{ c }{ Synthia-to-Cityscapes}&\\\midrule
         DAPS\cite{10203383}&73.7&26.5&71.9&1.0&0.0&7.6&9.9&12.4&81.4&77.4&27.4&23.1&\underline{47.0}&\underline{40.9}&12.6&\textbf{15.4}&64.7&42.2&33.0\\
         EDAPS\cite{edaps}&\underline{77.5}&\underline{36.9}&\underline{80.1}&\underline{17.2}&\textbf{1.8}&\underline{29.2}&\underline{33.5}&\underline{40.9}&\underline{82.6}&\underline{80.4}&\textbf{43.5}&\textbf{33.8}&45.6&35.6&\textbf{18.0}&2.8&\underline{72.7}&\underline{53.6}&\underline{41.2}\\
         Ours& \textbf{79.3}&\textbf{42.9}&\textbf{83.0}&\textbf{21.4}&\underline{0.6}&\textbf{33.7}&\textbf{40.8}&\textbf{52.5}&\textbf{83.2}&\textbf{83.8}&\underline{41.4}&\underline{30.1}&\textbf{48.1}&\textbf{42.8}&\underline{17.1}&\underline{4.2}&\textbf{74.1}&\textbf{56.6}&\textbf{44.1}\\\midrule
         &\multicolumn{19}{ c }{Synthia-to-Mapilliary}&\\\midrule
         CVRN\cite{Huang_2021_CVPR}&33.4&7.4&32.9&1.6&0.0&4.3&0.4&6.5&50.8&76.8&30.6&15.2&44.8&18.8&7.9&9.5&65.3&28.1&21.3\\
         EDAPS\cite{edaps}&\underline{77.5}&\underline{25.3}&\underline{59.9}&\underline{14.9}&\underline{0.0}&\underline{27.5}&\underline{33.1}&\underline{37.1}&\textbf{72.6}&\textbf{92.2}&\underline{32.9}&\underline{16.4}&\underline{47.5}&\underline{31.4}&\underline{13.9}&\underline{3.7}&\underline{71.7}&\underline{46.1}&\underline{36.6}\\
         Ours&\textbf{83.2}&\textbf{33.2}&\textbf{60.5}&\textbf{17.4}&\textbf{0.1}&\textbf{29.6}&\textbf{35.1}&\textbf{39.5}&\underline{71.8}&\underline{92.1}&\textbf{33.2}&\textbf{17.2}&\textbf{49.2}&\textbf{34.2}&\textbf{15.5}&\textbf{4.1}&\textbf{74.3}&\textbf{48.5}&\textbf{38.5}\\\bottomrule
         
    \end{tabular}}
    \caption{Per-class performance comparison with state-of-the-art UDA methods on panoptic segmentation for the Synthia-to-Cityscapes and Synthia-to-Mapilliary setups. Our proposal corresponds to the layer-wise model merging of MIC\cite{hoyer2023mic} for semantic segmentation and EDAPS\cite{edaps} for panoptic segmentation. Best results are indicated with bold while second best results are underlined.}
    \label{tab:SOTApan}
\end{table}

\paragraph{Merging of models on image classification and object detection}

\begin{table}[]
    \centering
    \addtolength{\tabcolsep}{-1pt}
    \resizebox{\textwidth}{!}{%
    \begin{tabular}{l c l c c}
        Task & Dataset  & Method& Reported & Ours\\\toprule
        Classification & Office-31 \cite{office31}& GLC \cite{GLC} &94.1 &94.8\\
        Classification & Office-Home \cite{officehome}& GLC \cite{GLC}&72.5 &74.2\\
        Classification & Office-Home \cite{officehome}& ConMix \cite{kumar2023conmix}&72.1 &73.2\\\midrule
        Detection & Cityscapes-to-Foggy Cityscapes \cite{SDV18}& MIC \cite{hoyer2023mic} & 47.6 & 48.1\\\bottomrule
    \end{tabular}}
    \caption{Layer-wise model merging on different tasks and UDA methods. Merging performed on checkpoints obtained during training. For classification we report average accuracy and for detection we report average precision.} 
    \label{tab:other tasks}
\end{table}

Table \ref{tab:other tasks} consolidates the outcomes derived from the application of layer-wise model merging across various UDA methods and tasks. Note that we employed the checkpoints acquired during the training phase for merging, thus ensuring that layer-wise weight merging incurs no additional computational overhead during neither training or evaluation.


\newpage

\section{Conclusions}
In this paper, we study the employment of model merging methods in the context of UDA to combine different models parameters. Additionally, we propose a method for merging models accounting for the specificity of the architecture layer depth. Our proposal can be easily included in other UDA frameworks as it only employs model checkpoints, or it can combine models using the same architecture but trained differently. Note that layer-wise model merging is computationally free in terms of training and inference. In a comprehensive evaluation, we have presented different comparatives to highlight the potential of model merging and the significant performance improvements our layer-wise merging achieves. For instance, improving the state-of-the-art performance by +2.3 and +2.9 (in terms of mIoU)on the Synthia-to-Cityscapes semantic and panoptic segmentation benchmarks respectively. We hope that, due to its high impact on performance and train-free nature, UDA researchers may introduce model merging into their frameworks.

\bmhead{Acknowledgment} This work has been partially supported by the Spanish Government through its TED2021-131643A-I00 (HVD) and the PID2021-125051OB-I00 (SEGA-CV) projects.

\section*{Declarations}
\bmhead{Conflict of Interests} The authors declare that they have no known competing financial interests or personal relationships that could have appeared
to influence the work reported in this paper.
\bmhead{Data availability}  The datasets analysed during the current work are publicly available in their respective repositories. The code will be publicly available upon acceptance in the LWMM repository: \href{http://www-vpu.eps.uam.es/publications/LWMM/}{http://www-vpu.eps.uam.es/publications/LWMM/}
%
%
\bibliography{main}
\end{document}